\documentclass{article}
\usepackage{arxiv}

\usepackage[utf8]{inputenc} 
\usepackage[T1]{fontenc}    
\usepackage{hyperref}       
\usepackage{url}            
\usepackage{booktabs}       
\usepackage{amsfonts}       
\usepackage{nicefrac}       
\usepackage{microtype}      
\usepackage{lipsum}         
\usepackage{graphicx}
\usepackage{doi}

\usepackage{amsmath,amssymb,amsfonts}
\usepackage{algorithmic}
\usepackage{graphicx}
\usepackage{textcomp}
\usepackage{microtype}
\usepackage{graphicx}
\usepackage{subfigure}
\usepackage{booktabs} 
\usepackage{longtable}
\usepackage{placeins}
\usepackage{psfrag}
\usepackage{algorithm}
\usepackage{array}
\usepackage[title]{appendix}%

\usepackage{amsmath}
\usepackage{amssymb}
\usepackage{mathtools}
\usepackage{amsthm}
\usepackage[table]{xcolor} 
\usepackage{lscape}
\usepackage{multirow}

\newcommand{\ApplyAbsoluteGradient}[1]{%
  \ifnum#1=1
    \cellcolor[RGB]{55,150,70}{} 
  \else\ifnum#1=2
    \cellcolor[RGB]{85,180,110}{} 
  \else\ifnum#1=3
    \cellcolor[RGB]{130,205,120}{} 
  \else\ifnum#1=4
    \cellcolor[RGB]{175,212,114}{} 
  \else\ifnum#1=5
    \cellcolor[RGB]{210,230,130}{} 
  \else  
    \cellcolor[RGB]{234,230,147}{} 
  \fi\fi\fi\fi\fi
}

\title{Polynomial Chaos Expanded Gaussian Process}

\newif\ifuniqueAffiliation
\uniqueAffiliationtrue

\usepackage{authblk}

\author{
  \begin{minipage}[t]{0.6\textwidth}
    \centering
    Dominik Polke*, Tim Kösters, Elmar Ahle \\
    Electrical Engineering and Computer Science \\
    University of Applied Sciences Niederrhein \\
    Krefeld, Germany\\
    \texttt{\{dominik.polke, elmar.ahle\}@hs-niederrhein.de}\\
    \texttt{tim.koesters@stud.hn.de}
  \end{minipage}
  \hfill
  \begin{minipage}[t]{0.3\textwidth}
    \centering
    Dirk Söffker \\
    Dynamics and Control \\
    University of Duisburg-Essen\\
    Duisburg, Germany \\
    \texttt{soeffker@uni-due.de}
  \end{minipage}
}

\begin{document}
\maketitle

\begin{abstract}
In complex and unknown processes, global models are initially generated over the entire experimental space but often fail to provide accurate predictions in local areas. A common approach is to use local models, which requires partitioning the experimental space and training multiple models, adding significant complexity. Recognizing this limitation, this study addresses the need for models that effectively represent both global and local experimental spaces. It introduces a novel machine learning (ML) approach: Polynomial Chaos Expanded Gaussian Process (PCEGP), leveraging polynomial chaos expansion (PCE) to calculate input-dependent hyperparameters of the Gaussian process (GP). This provides a mathematically interpretable approach that incorporates non-stationary covariance functions and heteroscedastic noise estimation to generate locally adapted models. The model performance is compared to different algorithms in benchmark tests for regression tasks. The results demonstrate low prediction errors of the PCEGP, highlighting model performance that is often competitive with or better than previous methods. A key advantage of the presented model is its interpretable hyperparameters along with training and prediction runtimes comparable to those of a GP.
\end{abstract}


\subsubsection*{Acknowledgement}
This work has been partially sponsored by the German Federal Ministry of Education and Research in the funding program "Forschung an Fachhochschulen", project I2DACH (grant no. 13FH557KX0, https://www.hs-niederrhein.de/i2dach). The authors would like to thank Fabio Sigrist for his help in applying the GPBoost algorithm~\cite{Sigrist.2022}. The authors acknowledge the use of ChatGPT (version 4, https://chat.openai.com) by OpenAI, for initial text refinement and stylistic improvements. The tool was employed to save time and enhance the quality of writing. All generated outputs were subsequently reviewed and corrected to ensure accuracy and alignment with the intended statements.
\newpage

\section{Introduction}
System modeling of real processes occurs within a specific global experimental space. A widely used choice for modeling unknown system behavior is the Gaussian process (GP), introduced in~\cite{Rasmussen.2006}. The GP is often used with stationary covariance functions for regression tasks. This assumes that the smoothness of a function is constant over the experimental space. As demonstrated in~\cite{Marmin.2018}, predicting a function with a varying smoothness using stationary covariance functions leads to increased prediction errors in areas with abrupt variations. To reduce the prediction errors, more experiments could be performed in the areas with abrupt variations, which leads to heterogeneous and non-uniform data point densities. According to~\cite{Marmin.2018}, stationarity assumes an almost uniform input design. When non-stationary covariance functions are used, advantages for data with non-stationarities are shown, e.g., in~\cite{Bitzer.2023} and~\cite{Marmin.2018}.


One approach to handle different densities of data across the entire design space with GPs is to use non-stationary covariance functions. Non-stationarity allows the modeling of the input-output relationship based on the input value range~\cite{Plagemann.2008}.
In~\cite{Paciorek.2003}, a non-stationary version of the Matérn covariance function is presented, which allows to model non-stationary behavior in data. A second level GP is used in~\cite{Plagemann.2008} to calculate the lengthscale parameters of the first level GP in order to model non-stationary behavior. For the anisotropic case, this leads to a second level GP for each input, which can be computationally expensive due to the high dimensions.
 
In~\cite{Hebbal.2021}, the use of deep Gaussian processes (DGPs), first introduced in~\cite{Damianou.2013}, with Bayesian optimization (BO) is investigated. The non-stationarity of data is modeled by considering a functional composition of stationary GPs, providing a multiple layer structure. The DGP approach has also shown promising results in learning surrogate models with as little data as possible, as presented in~\cite{Sauer.2023}. However, the DGP approach leads to a more difficult interpretability and is challenging for model training, due to the nested structure.

Furthermore, there are approaches, that use local experts to model non-stationary behavior, as presented in~\cite{Trapp.2020}. In~\cite{Bitzer.2023}, input partitioning with hierarchical hyperplane kernels (HHK) for GPs is used to achieve non-stationarity. Such approaches require training of several different models. 

In~\cite{Cremanns.2017}, the non-stationarity of GPs is achieved by using a deep neural network (DNN) for hyperparameter estimation of the GP depending on the input data. Typical DNNs are a type of black box learning, the training process of a DNN lacks transparency. This makes it difficult to understand how exactly a DNN is trained and how it generates decisions~\cite{Liang.2021}. 

These approaches result in either a structural limitation of the model, a computationally expensive model, the need for different local models, or a model that is difficult to interpret. In this work, a mathematically interpretable method for parameterization of non-stationary covariance functions and heteroscedastic noise estimation is presented. Here, the novel combination of polynomial chaos expansion (PCE) and Gaussian processes (GPs), called Polynomial Chaos Expanded Gaussian Process (PCEGP) is presented. The PCE is utilized to calculate input dependent hyperparameters of the GP. This approach provides the ability to model the global and local behavior of a complex process with only one model. 

The PCE offers several advantages, including low computational cost compared to nested model structures like DGP, since evaluating a polynomial function requires only basic arithmetic operations. As investigated in~\cite{Torre.2019}, the PCE also works well for machine learning (ML) regression on small datasets. In addition, the PCE representation allows the computation of statistical moments and sensitivities of the output with respect to the inputs, as described in~\cite{Wan.2020} and~\cite{Mara.2021}.

This paper is structured as follows. First, a short introduction and further references to PCE, GPs, and Bayesian hyperparameter optimization (HPO) are given in Section~\ref{background_and_related_work}. In Section~\ref{gppce_algorithm}, the ML approach with the novel combination of PCE and GP is introduced. Then, the method to generate non-stationary covariance functions and the heteroscedastic noise estimation are introduced. After that, the Bayesian HPO for the PCEGP is presented. The evaluation of the new ML approach is carried out in Section~\ref{experiments}, using various benchmark datasets. The results illustrate the PCEGP's ability to achieve low prediction errors across diverse applications, underscoring a model performance that competes with or surpasses previous methods. Finally, summary and outlook for future work are given in Section~\ref{summary_and_outlook}.

\newpage
\section{Background and Related Work}
\label{background_and_related_work}
In this section, the necessary components for the novel ML approach are presented. This includes the PCE, the GP, and the Bayesian optimization for hyperparameter optimization using the tree-structured Parzen estimator.

\subsection{Polynomial Chaos Expansion}
\label{pce_background}
The polynomial chaos expansion, presented in~\cite{Shen.2020}, is a popular method in uncertainty quantification (UQ) of models which are parametrized by independent random variables with extensions to dependent variables~\cite{Rahman.2018}. By viewing the model as an input-output map, the effect of input to output uncertainties can be assessed in UQ~\cite{Sudret.2017}. The PCE allows the modeling of  different data distributions with different kinds of polynomial bases.

In the case of UQ for high-fidelity models, a large number of simulations is required. To avoid the computational load, surrogate models can be used. One of the most widely used methods to build a surrogate model in UQ is the generalized polynomial chaos. The PCE is well suited for the approximation of functions with random variables, because of the orthogonality of the polynomial basis to the probability measure of the variables~\cite{Jakeman.2019}.

In general, the polynomial chaos expansion (PCE) is defined by
\begin{equation}
f\left(\boldsymbol{\xi}\right) \approx \sum_{\boldsymbol{k} \in \mathcal{K}} \alpha_{\boldsymbol{k}}\,\Phi_{\boldsymbol{k}}\left(\boldsymbol{\xi}\right),
\end{equation}
where $f\left(\boldsymbol{\xi}\right) \in \mathbb{R}$ denotes the approximated function by PCE, depending on the random variables $ \boldsymbol{\xi} = (\xi_1, \xi_2, \ldots, \xi_{n_\xi})$. The coefficients of the expansion are $\alpha_{\boldsymbol{k}}$, and the polynomial basis functions are $\Phi_{\boldsymbol{k}}(\boldsymbol{\xi})$. The multi-index $\boldsymbol{k} = (k_1, k_2, \ldots, k_{n_\xi})$ is a vector of indices, where $n_\xi$ denotes the number of random input variables, and $ \mathcal{K}$ is the set of all possible multi-indices.

The polynomial basis functions $\Phi_{\boldsymbol{k}}\left(\boldsymbol{\xi}\right)$ are products of univariate orthogonal polynomials $\phi_{k_i}\left(\xi_{i}\right)$ with $\xi_{i} \in \mathbb{R}$, defined as
\begin{equation}
\Phi_{\boldsymbol{k}}\left(\boldsymbol{\xi}\right) = \prod_{i=1}^{n_\xi} \phi_{k_i}\left(\xi_{i}\right).
\end{equation}
The orthogonality property of the polynomial basis functions $\Phi_{\boldsymbol{k}}\left(\boldsymbol{\xi}\right)$ with respect to the probability density function $p(\boldsymbol{\xi})$ of the random variables $\boldsymbol{\xi}$ can be expressed as
\begin{equation}
\int_{\Omega} \Phi_{\boldsymbol{k}}\left(\boldsymbol{\xi}\right) \, \Phi_{\boldsymbol{j}}\left(\boldsymbol{\xi}\right) \, p(\boldsymbol{\xi}) \, \mathrm{d}\boldsymbol{\xi} =
\begin{cases}
\gamma_{\boldsymbol{k}}, & \text{if } \boldsymbol{k} = \boldsymbol{j} \\
0, & \text{if } \boldsymbol{k} \neq \boldsymbol{j},
\end{cases}
\end{equation}
where $p(\boldsymbol{\xi})$ is the joint probability density function of the random variables, $\gamma_{\boldsymbol{k}}$ is a normalization constant associated with the multi-index $\boldsymbol{k}$, and $\Omega$ denotes the domain over which the random variables $\boldsymbol{\xi}$ are defined. This domain $\Omega$ encompasses all possible values of $\boldsymbol{\xi}$ for which the probability density function $p(\boldsymbol{\xi})$ is non-zero.

This condition can also be compactly represented using the Kronecker delta $\delta_{\boldsymbol{k}\boldsymbol{j}}$
\begin{equation}
\int_{\Omega} \Phi_{\boldsymbol{k}}\left(\boldsymbol{\xi}\right) \, \Phi_{\boldsymbol{j}}\left(\boldsymbol{\xi}\right) \, p(\boldsymbol{\xi}) \, \mathrm{d}\boldsymbol{\xi} = \gamma_{\boldsymbol{k}} \, \delta_{\boldsymbol{k} \boldsymbol{j}},
\end{equation}
where $\delta_{\boldsymbol{k}\boldsymbol{j}} = 1$ if $\boldsymbol{k} = \boldsymbol{j}$ and $0$ otherwise. 

The orthogonality of the basis functions ensures that the PCE provides a stable and efficient representation of the target function \( f(\boldsymbol{\xi}) \). This orthogonality simplifies the computation of the expansion coefficients \( \alpha_{\boldsymbol{k}} \) by ensuring that each coefficient can be determined independently of the others, reducing the complexity of the optimization process.

The summation in the general definition runs over all possible multi-indices $\boldsymbol{k} \in \mathcal{K}$, the expansion is generally infinite. To limit the expansion to a finite number of terms, the infinite summation is truncated to a finite sum with a maximum polynomial degree $q \in \mathbb{N}_0$

\begin{equation}
f\left(\boldsymbol{\xi}\right) \approx \sum_{\boldsymbol{k} \in \mathcal{K}_q} \alpha_{\boldsymbol{k}}\,\Phi_{\boldsymbol{k}}\left(\boldsymbol{\xi}\right),
\end{equation}
where $\mathcal{K}_q = \left\{ \boldsymbol{k} \in \mathcal{K} \, \big| \, \sum_{i=1}^{n_\xi} k_i \leq q \right\}$ is the set of multi-indices whose total degree $\sum_{i=1}^{n_\xi} k_i$ is less than or equal to the maximum degree $q$.

The choice of the polynomial basis function for each random variable \( \xi_{i} \) depends on the distribution of \( \xi_i \). There exist several orthogonal polynomial bases and their corresponding probability density functions. For example, the Legendre polynomials are defined by the polynomial function
\begin{equation}
\mathrm{Le}_n(\xi) = \frac{1}{2^n}\binom{2n}{n}(1-\xi^2)^{n/2}
\end{equation}
and the probability density function
\begin{equation}
p\left(\xi\right) = \frac{1}{2}.
\end{equation}

The number of coefficients $N_{\alpha}$ depends on the maximum polynomial degree $q$ and the number of random input variables $n_\xi$. It can be calculated by
\begin{equation}
N_{\alpha} = \binom{n_\xi + q}{q} = \frac{(n_\xi + q)!}{n_\xi! \, q!}.
\end{equation}
The number of coefficients $N_\alpha$ grows rapidly with increasing polynomial degree $q$ and the number of random input variables $n_\xi$. Therefore, it is crucial to keep the polynomial degree moderate or to reduce the number of random input variables using techniques such as principle component analysis (PCA)~\cite{Greenacre.2022}, to manage the rapid increase in the number of coefficients and mitigate the computational complexity of the expansion.

Another method to reduce the number of coefficients and the computational complexity is the application of sparse polynomial chaos expansion (SPCE), as presented in~\cite{Luthen.2021}. One way to achieve sparsity in the PCE is to apply Elastic Net (EN) regularization~\cite{Tian.2022}, which combines both L1~(Lasso) and L2~(Ridge) penalties. The Elastic Net optimization problem is defined as
\begin{align}
\min_{\alpha} &\left( \mathcal{L}(\alpha) + \lambda \left( 
\beta \sum_{\boldsymbol{k} \in \mathcal{K}} |\alpha_{\boldsymbol{k}}| \right)  
+ \lambda \left( (1 - \beta) \sum_{\boldsymbol{k} \in \mathcal{K}} \alpha_{\boldsymbol{k}}^2 \right) \right), \notag \\[-4mm]
&\hspace{17mm} \underbrace{\phantom{xxxxxx}}_{\text{L1 (Lasso) penalty}}  
\quad
\hspace{8mm} \underbrace{\phantom{xxxxx}}_{\text{L2 (Ridge) penalty}}
\end{align}
where $\mathcal{L}(\alpha)$ denotes the general loss function (e.g., mean squared error (MSE)), $\lambda \in \mathbb{R}^{+}$ is the regularization parameter that controls the overall strength of regularization, and $\beta \in [0, 1]$ is a parameter that determines the balance between the L1 and L2 penalties. This approach encourages sparsity while maintaining some of the stability properties of ridge regression. Moreover, this regularization technique helps to prevent overfitting, especially in cases where the polynomial degree is chosen too high, by penalizing overly complex models and reducing the variance of the estimated coefficients.

The application of PCE beyond its typical use case for UQ has been demonstrated in~\cite{Torre.2019} and~\cite{Schobi.2015}.  In~\cite{Torre.2019}, it is shown how the PCE can be used for ML regression. In~\cite{Schobi.2015}, a combination of PCE and GP for ML is presented, in which the PCE is used to describe the global behavior of the computational model while the GP describes the local variability of the model. This combination of PCE and GP is completely different to the approach developed in this work.

\subsection{Gaussian Process}
\label{gp_background}
The GP is a probabilistic ML approach using kernel machines~\cite{Rasmussen.2006}. The principle advantage over other kernel-based approaches is given by an uncertainty measurement of the predictions. 

A GP is a collection of random variables. Each finite number of these random variables have a joint Gaussian distribution. In the following the GP is considered for regression, where the training data are denoted as $\mathcal{D}=\{\boldsymbol{x}_i, y_i\}_{i=1}^{N}$ with $\boldsymbol{x}_i \in \mathbb{R}^{n_x}$ and $y_i \in \mathbb{R}$. The number of training points is denoted with $N$ and the number of inputs is denoted with~$n_x$.

The mean function $m(\boldsymbol{x})$ and the covariance function $k(\boldsymbol{x},\boldsymbol{x}')$, also called kernel function, completely describe the GP
\begin{equation}
f(\boldsymbol{x}) \sim \mathcal{GP}\left(m(\boldsymbol{x}){,}\,k(\boldsymbol{x},\boldsymbol{x}')\right),
\end{equation}
where $\boldsymbol{x}$ and $\boldsymbol{x}'$ represent input data points. Most ML applications have a large number of input parameters, e.g., in chemical engineering~\cite{Li.2021}, autonomous driving~\cite{Kiran.2022}, and energy consumption forecasting~\cite{Somu.2021}. This results in a high-dimensional feature space for the learning algorithm. For distance-based ML algorithms, this leads to the so-called curse of dimensionality~\cite{Keogh.2011}. To avoid this, unimportant features can be removed with automatic relevance determination (ARD)~\cite{Neal.1996} or by techniques, presented in~\cite{Varunram.2021}. For GPs, the ARD is realized by introducing a lengthscale parameter for each feature~\cite{Paananen.2019}.

A common choice for the stationary covariance function of the GP is the squared exponential kernel, see~\cite{Rasmussen.2006}. It is defined by
\begin{equation}
k\left(\boldsymbol{x},\boldsymbol{x}'\right) = \sigma_{\mathrm{f}}^2\,\exp\left(-\frac{\|\boldsymbol{x}-\boldsymbol{x}'\|^2}{2\,l^2}\right),
\end{equation}
with the correlation lengthscale $l \in \mathbb{R}^{+}$ and the signal variance $\sigma_{\mathrm{f}}^2 \in \mathbb{R}^{+}$. The noise variance $\sigma^2_{\mathrm{n}} \in \mathbb{R}^{+}$ is used to model the noise in the data and also serves as hyperparameter of the GP. There exist also extensions of this covariance function, e.g., the squared exponential covariance function with ARD, which is defined by
\begin{equation}
k\left(\boldsymbol{x},\boldsymbol{x}'\right) = \sigma_{\mathrm{f}}^2\,\exp\left(-\frac{1}{2}\left(\boldsymbol{x}-\boldsymbol{x}'\right)^\intercal\boldsymbol{L}\left(\boldsymbol{x}-\boldsymbol{x}'\right)\right),
\end{equation}
with the lengthscale matrix $\boldsymbol{L}= \mathrm{diag}\left(\left[l_1 \dots l_{n_x}\right]\right)^{-2}$. The lengthscale matrix includes one lengthscale parameter $l_i \in \mathbb{R}^{+}$ for each input. A further extension is presented in~\cite{Heinonen.2016}, where the authors used a non-stationary generalization of the squared exponential covariance function, defined as
\begin{equation}
k\left(\boldsymbol{x}, \boldsymbol{x}'\right) = \sigma_\mathrm{f}\left(\boldsymbol{x}\right) \sigma_\mathrm{f}\left(\boldsymbol{x}'\right) \sqrt{\frac{2\,l\left(\boldsymbol{x}\right) l \left(\boldsymbol{x}'\right)}{l^2\left(\boldsymbol{x}\right) + l^2\left(\boldsymbol{x}'\right)}} \mathrm{exp} \left(- \frac{\|\boldsymbol{x}-\boldsymbol{x}'\|^2}{l^2\left(\boldsymbol{x}\right) + l^2\left(\boldsymbol{x}'\right)} \right).
\end{equation}
In this extension, the hyperparameters $\sigma_\mathrm{f}\left(\boldsymbol{x}\right) \in \mathbb{R}^{+}$ and $l\left(\boldsymbol{x}\right) \in \mathbb{R}^{+}$ are data point dependent to model non-stationarity.

The hyperparameters are used to fit the model to the given data. In the following, all hyperparameters of the GP are summarized in $\boldsymbol{\theta} \in \mathbb{R}^{n_x+2}$, which leads to the covariance matrix
\begin{equation}
\boldsymbol{K}\left(\boldsymbol{\boldsymbol{X}, \theta}\right)= \boldsymbol{K}\left(\boldsymbol{X}, \boldsymbol{L}, \sigma_{\mathrm{f}}\right)+\sigma^2_{\mathrm{n}}\,\boldsymbol{I}.
\end{equation}

In the following, $\boldsymbol{K}\left(\boldsymbol{X}\right) \in \mathbb{R}^{N \times N}$ denotes the covariance matrix depending on the training data and $\boldsymbol{k}\left(\boldsymbol{X}, \boldsymbol{x}^*\right) \in \mathbb{R}^{N}$ the covariance vector depending on the training data and the next point $\boldsymbol{x}^*$ to predict.  The input training points of the whole dataset~$\mathcal{D}$ are denoted as $\boldsymbol{X} \in \mathbb{R}^{N \times n_x}$. The elements of the covariance matrix $\boldsymbol{K}\left(\boldsymbol{X}\right)$ are calculated by the covariance function $k\left(\boldsymbol{x}, \boldsymbol{x}'\right)$ and thus dependent on the training data~$\mathcal{D}$ and the hyperparameters $\boldsymbol{L}$, $\sigma_\mathrm{f}$, and $\sigma_{\mathrm{n}}$. 

Usually, these hyperparameters are optimized by maximizing the log marginal likelihood (LML), which is given by
\begin{align}
\mathrm{log}\,p\left(\boldsymbol{y}|\,\boldsymbol{X} \right) = -\frac{1}{2}\,\boldsymbol{y}^\intercal \boldsymbol{K}\left(\boldsymbol{X}\right)^{-1}\boldsymbol{y}\nonumber -\frac{1}{2}\,\mathrm{log}|\boldsymbol{K}\left(\boldsymbol{X}\right)| - \frac{N}{2}\,\mathrm{log}\left(2\pi\right)
\end{align}
under the assumption that $\boldsymbol{y} \sim \mathcal{N}\left( 0, \boldsymbol{K}\left(\boldsymbol{X}\right) + \sigma^2_{\mathrm{n}}\,\boldsymbol{I}\right)$ for the GP output. For the optimization of the hyperparameters with a gradient descent algorithm, the gradient of the LML is used and given by
\begin{align}
\label{eq_gradient_maximum_likelihood}
\frac{\partial}{\partial \boldsymbol{\theta}_i} \mathrm{log}\,p\left(\boldsymbol{y}|\,\boldsymbol{X}, \boldsymbol{\theta} \right) = \frac{1}{2}\,\boldsymbol{y}^{\intercal}\,\boldsymbol{K}\left(\boldsymbol{X}\right)^{-1} \frac{\partial}{\partial \boldsymbol{\theta}_i} \boldsymbol{K}\left(\boldsymbol{X}\right)^{-1} \boldsymbol{y} - \frac{1}{2} \mathrm{tr} \left(\boldsymbol{K}\left(\boldsymbol{X}\right)^{-1}\frac{\partial \boldsymbol{K}\left(\boldsymbol{X}\right)}{\partial \boldsymbol{\theta}_i} \right).
\end{align}
To predict an output based on a new input $\boldsymbol{x}^* \in \mathbb{R}^{n_x}$ and the given dataset $\mathcal{D}$, the posterior distribution is defined by
\begin{align}
\mu\left(\boldsymbol{x}^*\right) =&\, \boldsymbol{k}\left(\boldsymbol{X},\boldsymbol{x}^* \right)^\intercal  \boldsymbol{K} \left(\boldsymbol{X}\right)^{-1} \boldsymbol{y}, \\
\sigma^2\left(\boldsymbol{x}^*\right) =& \, k\left(\boldsymbol{x}^*,\boldsymbol{x}^* \right) - \boldsymbol{k}\left(\boldsymbol{X},\boldsymbol{x}^* \right)^{\intercal} \boldsymbol{K}\left(\boldsymbol{X}\right)^{-1} \boldsymbol{k}\left(\boldsymbol{X},\boldsymbol{x}^* \right),
\end{align}
where $\mu\left(\boldsymbol{x}^*\right) \in \mathbb{R}$ denotes the mean prediction and $\sigma^2\left(\boldsymbol{x}^*\right) \in \mathbb{R}^{+}$ the uncertainty prediction of the point to evaluate $\boldsymbol{x}^*$.


\subsection{Bayesian Hyperparameter Optimization Using the Tree-Structured Parzen Estimator\label{bo_background}}
Bayesian optimization (BO) is a powerful method for efficient global optimization of expensive and unknown black-box functions~\cite{Greenhill.2020}. In the specific case of HPO with BO, the goal is to find an optimal set of hyperparameters $\boldsymbol{h}_{\mathrm{opt}}$ that minimizes a given objective function $o_\mathrm{B}(\boldsymbol{h})$, which represents the performance of a machine learning model. This can be formally stated as
\begin{equation}
\boldsymbol{h}_{\mathrm{opt}} \in \operatornamewithlimits{argmin}_{\boldsymbol{h} \in \mathcal{H}} o_\mathrm{B}(\boldsymbol{h}),
\end{equation}
where $\mathcal{H}$ denotes the space of all possible hyperparameter configurations, with $\boldsymbol{h}~\in~\mathbb{R}^{N_{\mathrm{HP}}}$ representing a vector of $N_{\mathrm{HP}}$ hyperparameters. For ML models, the objective function $o_\mathrm{B}(h) \in \mathbb{R}$ often involves minimizing metrics such as the error rate in classification tasks or the MSE in regression tasks.

In the context of BO, the observations from previous evaluations are represented as $\mathcal{D}_{\mathrm{HP}} = \{(\boldsymbol{h}_i, o_{\mathrm{B},i})\}_{i=1}^{N_{\mathrm{obs}}}$, where $h_i \in \mathbb{R}^{N_{\mathrm{HP}}}$ are the hyperparameters and $o_{\mathrm{B},i} \in \mathbb{R}$ are the corresponding objective values. The total number of observations is denoted by $N_{\mathrm{obs}}$.

The BO achieves the minimization of the error metric by iteratively constructing a probabilistic surrogate model to approximate the objective function and then using an acquisition function to decide where to sample next. The acquisition function balances exploration, which searches less-explored areas for potentially better solutions, and exploitation, which focuses on optimizing known promising regions. As new data are collected, the model is updated to improve its accuracy and guide the search process more effectively~\cite{Greenhill.2020}. A commonly choice of the acquisition function is the expected improvement (EI) acquisition function. This function balances exploration with exploitation during the search~\cite{White.2021}. The EI acquisition function is described by
\begin{align}
\mathrm{EI} \left(\boldsymbol{h},\,\rho\right) =& \left(\mu - f\left(\boldsymbol{h}^*\right)- \rho\right) p \left(\frac{\mu - f\left(\boldsymbol{h}^*\right)- \rho}{\sigma} \right) + \sigma\,\varphi \left(\frac{\mu - f\left(\boldsymbol{h}^*\right)- \rho}{\sigma} \right),
\end{align} 
with the mean $\mu \in \mathbb{R}$, the standard deviation $\sigma \in \mathbb{R}^{+}$, the so far observed minima $f\left(x^*\right) \in \mathbb{R}$, and a weight factor $\rho \in \mathbb{R}$ for fine tuning between exploitation and exploration. In addition, $p$ denotes the probability density function and $\varphi$ denotes the cumulative distribution function.
A powerful method within BO for HPO is the tree-structured Parzen estimator (TPE). The TPE leverages kernel density estimators (KDEs) to model the probability distribution $p(h|o_\mathrm{B}, \mathcal{D}_{\mathrm{HP}})$ of the observed data points~\cite{Ozaki.2020}. In each iteration, the TPE algorithm splits the data into two groups: a "better group" ($\mathcal{D}_{\mathrm{HP,b}}$), which includes hyperparameters yielding better objective values, and a "worse group" ($\mathcal{D}_{\mathrm{HP,w}}$), which consists of hyperparameters with inferior objective values. 
The specific strategy for splitting the data into $\mathcal{D}_{\mathrm{HP,b}}$ and $\mathcal{D}_{\mathrm{HP,w}}$ is influenced by the choice of the quantile $\vartheta \in (0, 1]$. This quantile is dynamically adjusted based on the number of observations $N_{\mathrm{obs}}$. Two common strategies for defining $\vartheta$ are
\begin{align}
\text{Linear:} \quad & \vartheta = \psi_1, \quad \psi_1 \in (0, 1], \\
\text{Square-root:} \quad & \vartheta = \frac{\psi_2}{\sqrt{N_{\mathrm{obs}}}}, \quad \psi_2 \in (0, \sqrt{N_{\mathrm{obs}}}]\,.
\end{align}

Here, $\psi_1$ and $\psi_2$ are parameters that control the balance between exploration and exploitation. The linear approach, with a fixed quantile $\vartheta$, tends to focus on a constant fraction of the best-performing observations, promoting exploitation. Conversely, the square-root strategy dynamically adjusts $\vartheta$ as more data are acquired, favoring exploration by increasing the search in less-explored regions~\cite{Bergstra.2011}. 

\begin{figure}[htbp]
    \centering
    \includegraphics[width=\textwidth]{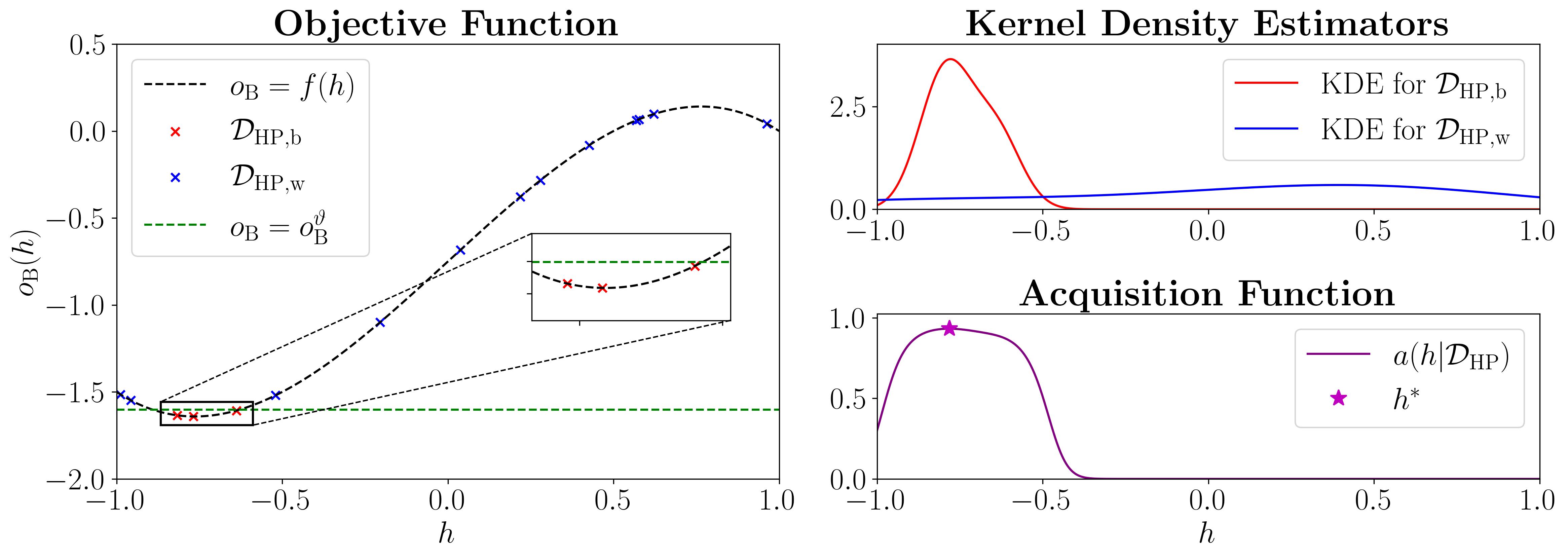}
    \caption{
        Visualization of the tree-structured Parzen estimator (TPE) process for Bayesian hyperparameter optimization}
    \label{fig:tpe_visualization}
\end{figure}

In Figure~\ref{fig:tpe_visualization}, the process of Bayesian HPO with TPE is illustrated. The left plot depicts the objective function $o_\mathrm{B} = f(h)$ along with its observations. A chosen quantile $\vartheta$ defines the boundary that separates the "better group" $\mathcal{D}_{\mathrm{HP,b}}$ from the "worse group" $\mathcal{D}_{\mathrm{HP,w}}$. The top right plot shows the corresponding KDEs, representing the distributions of the better group $\mathcal{D}_{\mathrm{HP,b}}$ and the worse group $\mathcal{D}_{\mathrm{HP,w}}$. The bottom right plot visualizes the acquisition function $a(h | \mathcal{D}_{\mathrm{HP}})$, which is computed as the ratio of these KDEs. The star marks the next hyperparameter evaluation point, corresponding to the maximum of the acquisition function.
\newpage
To describe this approach mathematically, TPE models the conditional probability as follows

\begin{align}
p(h|o_\mathrm{B}, \mathcal{D}_{\mathrm{HP}}) = 
\begin{cases}
p(\boldsymbol{h} | \mathcal{D}_{\mathrm{HP,b}}) & \text{if } o_\mathrm{B} \leq o_\mathrm{B}^\vartheta \\
p(\boldsymbol{h} | \mathcal{D}_{\mathrm{HP,w}}) & \text{if } o_\mathrm{B} > o_\mathrm{B}^\vartheta\,.
\end{cases} 
\end{align}

In this formulation, $\mathcal{D}_{\mathrm{HP,b}}$ contains the observations where the objective value is within the top $\vartheta$-quantile, while $\mathcal{D}_{\mathrm{HP,w}}$ includes those with objective values outside this quantile. The KDEs for these groups are estimated as

\begin{align}
p(\boldsymbol{h}|\mathcal{D}_{\mathrm{HP,b}}) &= w_{0,\mathrm{b}} p_0(h) + \sum_{i=1}^{N_{\mathrm{b}}} w_{i,\mathrm{b}} k_{\mathrm{KDE}}(h, h_i | b_{\mathrm{b}}), \\
p(\boldsymbol{h}|\mathcal{D}_{\mathrm{HP,w}}) &= w_{0,\mathrm{w}} p_0(h) + \sum_{i=N_{\mathrm{b}}+1}^{N} w_{i,\mathrm{w}} k_{\mathrm{KDE}}(h, h_i | b_\mathrm{w}),
\end{align}

where $w_{i,\mathrm{b}}, w_{i,\mathrm{w}} \in [0,1]$ are weights, $k_{\mathrm{KDE}}$ is the kernel density function used for KDE, and $b_\mathrm{b}, b_\mathrm{w} \in \mathbb{R}^+$ are the bandwidths for the KDEs. The acquisition function used by TPE is derived from the ratio of these densities

\begin{equation}
a(h | \mathcal{D}_{\mathrm{HP}}) = \frac{p(h | \mathcal{D}_{\mathrm{HP,b}})}{p(h | \mathcal{D}_{\mathrm{HP,w}})}.
\end{equation}

This acquisition function guides the selection of the next set of hyperparameters by favoring those regions where the likelihood of improvement is higher, balancing the trade-off between exploration and exploitation similar to the expected improvement~(EI) function~\cite{White.2021}.

A stopping criterion, such as a maximum number of iterations or convergence, determines how long BO for HPO continues iteratively. This method is especially useful for complex search spaces, where the evaluation of the objective function is costly~\cite{Wu.2019, Nguyen.2019, Victoria.2021}.

Several factors may contribute to the advantages of the TPE over GPs in HPO, as investigated in~\cite{Bergstra.2011}. A key advantage is its ability to naturally balance exploration and exploitation, particularly in search spaces with complex, discrete, or categorical variables. While both methods are effective in different contexts, TPE's flexibility and robustness in handling diverse search spaces make it a potentially more suitable choice in certain scenarios~\cite{Bergstra.2011}.

\section{Polynomial Chaos Expanded Gaussian Processes}\label{gppce_algorithm}
In this section, the novel ML approach as a combination of PCE and GP is presented, where the PCE is used to calculate the hyperparameters of the GP. First, the general architecture and methodology of this approach is presented. After that, the hyperparameter optimization is described. Then, a runtime comparison is carried out. 

\subsection{Architecture and Methodology}
The general architecture of the PCEGP is shown in Figure~\ref{fig_pcegp_framework}. It consists of data scaling of the input~$\boldsymbol{x}^*$, data point dependent hyperparameter calculation, and the prediction of the output $\hat{\boldsymbol{y}}_{\mathrm{PCEGP}}$ in $\mathbb{R}^{2}$, which includes the mean and uncertainty prediction. First, the given input data point $\boldsymbol{x}^*$ is scaled to $\boldsymbol{x}_\mathrm{s}^*$ with a scaler, e.g., the min-max-scaler.

\begin{figure*}[b!]
  \centering
  \includegraphics[width=\textwidth]{./Figure2.eps} 
  \caption{Evaluation of the Polynomial Chaos Expanded Gaussian Process with data scaling, point-dependent hyperparameter calculation, hyperparameter transformation, and output prediction}
  \label{fig_pcegp_framework}
\end{figure*}

As illustrated in Figure~\ref{fig_pcegp_framework}, the scaled data point $\boldsymbol{x}_\mathrm{s}^*$ is the input for the PCEs used to predict the hyperparameters $\hat{l}_1\left(\boldsymbol{x}_\mathrm{s}^*\right) \dots \hat{l}_{n_k}\left(\boldsymbol{x}_\mathrm{s}^*\right)$ for each covariance function $k_1 \dots k_{n_k}$, where $n_k \in \mathbb{N}$ denotes the number of used covariance functions. The scaled data point is also the input for the PCE used to calculate the heteroscedastic noise parameter $\hat{\sigma}_{\mathrm{n}}\left(\boldsymbol{x}_{\mathrm{s}}^{*}\right)$. The lengthscale parameters $\hat{l}_i\left(\boldsymbol{x}_{\mathrm{s}}^{*}\right)$ predicted by the PCE are transformed to ensure that the lengthscale $\hat{l}_{i,\mathcal{T}}\left(\boldsymbol{x}_{\mathrm{s}}^{*}\right)$ is $\in \mathbb{R}^{+}$. Similarly, the predicted noise $\hat{\sigma}_{\mathrm{n}}\left(\boldsymbol{x}_{\mathrm{s}}^{*}\right)$ is transformed to $\hat{\sigma}_{\mathrm{n},\mathcal{T}}\left(\boldsymbol{x}_{\mathrm{s}}^{*}\right)$ to ensure it remains in $\mathbb{R}^{+}$.

The choice of the PCE basis function depends on the input data. In Appendix~\ref{appendix_pce_bases}, different polynomial bases are shown. The approach of this study allows the combination of different polynomial bases for the calculation of the GP hyperparameters depending on the input data. This eliminates the choice of the polynomial basis, and the weighting of these becomes part of the HPO. As shown in Figure~\ref{fig_pcegp_framework}, the combination of different PCE basis functions is achieved by a linear combination of these bases. 

Non-stationary covariance functions enable the model to dynamically adjust to functions characterized by varying degrees of smoothness and training point densities across the input space. For GPs, this can be achieved by the estimation of an input dependent lengthscale parameter $l\left(\boldsymbol{x}\right)$ of a chosen covariance function. In this work, PCE is used to calculate the lengthscale parameters as a function of the input data.

For this purpose, the stationary covariance functions, have to be adapted to estimate a data point dependent lengthscale $l\left(\boldsymbol{x}\right)$. For example, the squared exponential covariance function 
\begin{equation}
k\left( \boldsymbol{x},\boldsymbol{x}'\right) = \sigma_\mathrm{f}^2\,  \mathrm{exp}\left( -\frac{\| \boldsymbol{x}-\boldsymbol{x}'\|^2}{2\,l^2}\right),
\end{equation}
is changed to
\begin{equation}
k\left( \boldsymbol{x},\boldsymbol{x}'\right) = \sigma_\mathrm{f}^2\,  \mathrm{exp}\left( -\frac{\| l\left(\boldsymbol{x}\right)\, \boldsymbol{x}-l \left(\boldsymbol{x}'\right)\,\boldsymbol{x}'\|^2}{2}\right),
\end{equation}
which enables the possibility to calculate the lengthscale for each data point separately by PCE. This approach is adapted from~\cite{Cremanns.2017} and~\cite{Cremanns.2021}, where this is done via a DNN. This makes it possible to transform any stationary covariance function into a non-stationary one. Here, $l\left(\boldsymbol{x}\right)$ is generally described as an input-dependent lengthscale. The lengthscale value calculated by PCE is denoted with $\hat{l}\left(\boldsymbol{x}\right)$. In Figure~\ref{fig_pcegp_framework}, the case $\hat{l}\left(\boldsymbol{x}_\mathrm{s}^*\right)$ is shown, where the data point $\boldsymbol{x}_\mathrm{s}^*$ is already scaled. 

To avoid problems with the covariance calculation, an output scaling by standardization is used in the following, which also corresponds to the previously mentioned assumption of $\boldsymbol{y} \sim \mathcal{N}\left( 0, \boldsymbol{K}\left(\boldsymbol{X}\right) + \sigma^2_{\mathrm{n}}\,\boldsymbol{I}\right)$ for the GP output.

The choice of the polynomial basis depends on the distribution of the input data. With
\begin{equation}
\label{eq_lengthscale_pce}
\hat{l}\left(\boldsymbol{x}\right) = \sum_{\boldsymbol{k} \in \mathcal{K}_l} \alpha_{l,\boldsymbol{k}}\,\Phi_{\boldsymbol{k}}\left(\boldsymbol{x}\right), 
\end{equation}
the lengthscale becomes an input dependent hyperparameter for the covariance functions, estimated by the PCE. From this point onward, the optimizer's task shifts from directly adjusting the lengthscale parameters to optimizing the $N_{\alpha,l} \in \mathbb{N}$ coefficients of the PCE for lengthscale calculation. In the case where the lengthscale parameter is calculated by a linear combination of $b$ polynomial expansions, the number of parameters to optimize changes to 
\begin{equation}
N_{\alpha,l} = b\,\binom{n_x + q_l}{q_l} = b\,\frac{(n_x + q_l)!}{n_x! \, q_l!},
\end{equation}
where $q_l$ denotes the polynomial degree of the PCE for lengthscale estimation. The polynomial coefficients of the input dependent lengthscale $\hat{l}\left(\boldsymbol{x}\right)$ are summarized in the vector $\boldsymbol{\alpha}_l \in \mathbb{R}^{N_{\alpha,l}}$.

The superposition of the polynomial expansions is achieved by a linear combination of the polynomials, as demonstrated in Figure~\ref{fig_pcegp_framework} with 
\begin{align}
\hat{l}\left(\boldsymbol{x}\right) =  \sum_{\boldsymbol{k} \in \mathcal{K}_l} \alpha_{l,\boldsymbol{k},1}\,\Phi_{\boldsymbol{k},1}\left(\boldsymbol{x}\right) + \sum_{\boldsymbol{k} \in \mathcal{K}_l} \alpha_{l,\boldsymbol{k},2}\,\Phi_{\boldsymbol{k},2}\left(\boldsymbol{x}\right) + \dots + \sum_{\boldsymbol{k} \in \mathcal{K}_l} \alpha_{l,\boldsymbol{k},b}\,\Phi_{\boldsymbol{k},b}\left(\boldsymbol{x}\right).
\end{align}

A separate PCE is also used to model the heteroscedastic noise parameter $\hat{\sigma}_\mathrm{n}^2\left(\boldsymbol{x}\right)$ of the GP as a function of the input data. The estimation of the noise parameter is defined by
\begin{equation}
\label{eq_noise_pce}
\hat{\sigma}_{\mathrm{n}}^2\left(\boldsymbol{x}\right) = \sum_{\boldsymbol{k} \in \mathcal{K}_{\sigma_\mathrm{n}}} \alpha_{\sigma_\mathrm{n},\boldsymbol{k}}\,\Phi_{\boldsymbol{k}}\left(\boldsymbol{x}\right).
\end{equation}
As the noise of the signal can be dependent on the measurement range for real sensor data but often does not reflect a complex function, truncation of the polynomial degree is possible to a significantly smaller value $q_{\sigma_\mathrm{n}}$ than for the lengthscale calculation. This also limits the number of parameters $N_{\alpha,\sigma_\mathrm{n}} \in \mathbb{N}$ to be optimized for modeling the noise variance 
\begin{equation}
N_{\alpha,\sigma_\mathrm{n}} =  \binom{n_x + q_{\sigma_\mathrm{n}}}{q_{\sigma_\mathrm{n}}} =  \frac{(n_x + q_{\sigma_\mathrm{n}})!}{n_x! \, q_{\sigma_\mathrm{n}}!}.
\end{equation}
The coefficients for noise estimation by PCE are summarized in $\boldsymbol{\alpha}_{\sigma_\mathrm{n}} \in \mathbb{R}^{N_{\alpha,\sigma_\mathrm{n}}}$.

In GP models, the lengthscale and noise parameters are critical in defining the covariance function that determines the covariance between input points. The lengthscale controls how quickly the correlation between data points decays with increasing distance, while the noise parameter represents the variance of the observed noise. For these hyperparameters to function correctly, they must satisfy certain constraints to ensure a valid covariance structure.

Ensuring that the lengthscale parameter $\hat{l}(\boldsymbol{x})$ remains positive is essential, as a non-positive lengthscale would violate the mathematical requirements for the GP. If the lengthscale is zero or negative, the covariance function may not properly represent the distance-dependent correlations between points, resulting in an invalid covariance matrix. A positive lengthscale ensures that the kernel function decreases smoothly as the distance between points increases, which is necessary for the model to behave predictably and for the covariance matrix to remain positive semidefinite (PSD). This requirement means that for any trivial vector $\boldsymbol{v} \in \mathbb{R}^{n_x}$, the following condition must hold:

\begin{equation}
\boldsymbol{v}^{\intercal} \boldsymbol{K}(\boldsymbol{X})\, \boldsymbol{v} \geq 0.
\end{equation}

Similarly, the noise variance $\hat{\sigma}_{\mathrm{n}}^2(\boldsymbol{x})$ must be non-negative. The noise variance contributes to the diagonal elements of the covariance matrix. If $\hat{\sigma}_{\mathrm{n}}^2(\boldsymbol{x}) < 0$, the diagonal elements of the covariance matrix may become negative, which would violate the requirement for positive semidefiniteness. Negative noise variances do not make sense in a probabilistic model, as variances cannot be negative. Thus, ensuring that $\hat{\sigma}_{\mathrm{n}}^2(\boldsymbol{x}) \geq 0$ is crucial for maintaining the validity of the GP.

When using methods such as the PCE to estimate the lengthscale and noise variance of a GP, it is possible that the values generated may be zero or negative. The PCE can generate any real-valued output within its stochastic domain, including values that are not appropriate for parameters such as lengthscale and noise. A transformation is required to ensure that these hyperparameters satisfy the necessary constraints.

A transformation $\mathcal{T}$ is applied to map the estimated lengthscale $\hat{l}(\boldsymbol{x})$ and noise $\hat{\sigma}_{\mathrm{n}}^2(\boldsymbol{x})$ to valid domains:

\begin{equation}
\hat{l}_\mathcal{T}\left(\boldsymbol{x}\right) = \mathcal{T}\left(\hat{l}\left(\boldsymbol{x}\right)\right), \quad \hat{\sigma}_{\mathrm{n},\mathcal{T}}^2 \left(\boldsymbol{x}\right) = \mathcal{T}\left(\hat{\sigma}_{\mathrm{n}}^2\left(\boldsymbol{x}\right)\right).
\end{equation}

A common choice for such a transformation is the softplus transformation, which maps any real number to $\mathbb{R}^{+}$. The softplus transformation is defined as

\begin{equation}
\hat{l}_\mathcal{T}\left(\boldsymbol{x}\right) = \mathrm{log}\left(1+\mathrm{e}^{\hat{l}\left(\boldsymbol{x}\right)}\right) \quad \text{and} \quad \hat{\sigma}_{\mathrm{n},\mathcal{T}}^2 = \mathrm{log}\left(1+\mathrm{e}^{\hat{\sigma}_{\mathrm{n}}^2\left(\boldsymbol{x}\right)} \right).
\end{equation}
This transformation ensures that the transformed lengthscale $\hat{l}_\mathcal{T}(\boldsymbol{x})$ and noise $\hat{\sigma}_{\mathrm{n},\mathcal{T}}^2(\boldsymbol{x})$ are strictly within $\mathbb{R}^{+}$. By applying this approach, both the lengthscale and the noise remain positive during the optimization process, preserving the mathematical properties required for the GP model to be valid and numerically stable. This ensures that the covariance matrix remains PSD and that the hyperparameters adhere to the required constraints. Different commonly used transformations are shown in Appendix~\ref{appendix_pce_transformations}.

In order to perform versatile modeling tasks with this novel approach, different covariance functions are combined and the influence of the respective covariance function is controlled by HPO. The covariance functions used in this work are shown in Appendix~\ref{appendix_covariance_functions} and can be exchanged or extended by further ones as needed. The combination of the covariance functions is realized by summation of the covariance functions with
\begin{equation}
k_\mathrm{\Sigma}\left(\boldsymbol{x},\boldsymbol{x}'\right) = k_1\left(\boldsymbol{x},\boldsymbol{x}'\right) + k_2\left(\boldsymbol{x},\boldsymbol{x}'\right) + \dots + k_{n_k}\left(\boldsymbol{x},\boldsymbol{x}'\right).
\end{equation}
The covariance matrix is then computed from the combined covariance function, and the prediction $\hat{\boldsymbol{y}}_\mathrm{PCEGP,s}$ is determined by combining this matrix with the estimated noise. Finally, the predictions are rescaled to the original range $\hat{\boldsymbol{y}}_\mathrm{PCEGP}$. In Appendix~\ref{appendix_model_prediction}, the algorithm for model prediction is shown. In this work, the covariance functions and polynomial bases are chosen manually and are not part of the HPO. In Section~\ref{experiments}, the setup of the model for the experiments carried out in this work is presented in detail

\subsection{Hyperparameter Optimization}\label{hpo}
In this work, the hyperparameter optimization is done via the tree-structured Parzen estimator and the Nadam optimizer, which is introduced in~\cite{dozat2016nadam} and combines the adaptive learning rate strategy of Adam~\cite{Kingma.2015} with the Nestorov momentum. The Nadam optimizer is well-suited for the PCEGP approach as it efficiently handles the non-convexity and high dimensionality associated with optimizing PCE coefficients for non-stationary lengthscale calculations in GPs. The optimization loop is described in Appendix~\ref{appendix_hpo}. The first step is to investigate the data for the choice of a suitable input and output scaler, covariance functions, and polynomial bases. After that, the data are scaled. A termination criterion has to be selected for the optimization loop, i.e., the number of trials for hyperparameter optimization. Then, the hyperparameter optimization starts with the TPE as a Bayesian approach. First, initial hyperparameter combinations are placed with random search in the entire hyperparameter space. Then, the hyperparameters are suggested via the TPE algorithm. 

The TPE algorithm uses an objective function for hyperparameter optimization. The inputs of this function are the hyperparameters of the model. This includes the selection of the polynomial degree $q_l$ for the lengthscale $\hat{l}\left(\boldsymbol{x}\right)$ and the polynomial degree $q_{\sigma_{\mathrm{n}}}$ for the heteroscedastic noise variance $\hat{\sigma}_\mathrm{n}^2\left(\boldsymbol{x} \right)$ calculation. In addition, the scaling factors $\boldsymbol{\epsilon}$, the learning rate $\eta$, the number of gradient iterations $n_\mathrm{I}$, and the initial values of the PCEs coefficients $c_0$ are suggested via the TPE. In the case that elastic net regularization is used, the regularization parameters $\beta$ and $\lambda$ are also used as hyperparameters for the BO.

The output of the objective function is the loss of the model. The optimization objective is to minimize the loss metric. For this, a shuffled $k$-fold cross-validation is performed, in which $k$ different models are trained using a gradient descent method and the hyperparameters of the PCEGP are fine-tuned. The gradient is built over $n_{\mathrm{I}}$ iterations for hyperparameter fine-tuning.

The use of the TPE ensures that the gradient descent procedure is performed with different hyperparameter combinations to find the global optimum. Compared to the grid search or random search, the Bayesian HPO with TPE has the advantage that it learns from poor hyperparameter combinations and avoids these regions for new hyperparameter suggestions. 

For the gradient descent method of model hyperparameter optimization, the covariance matrix depends now on the lengthscale PCE coefficients $\boldsymbol{\alpha}_{l}$ and the noise PCE coefficients $\boldsymbol{\alpha}_{\sigma_\mathrm{n}}$, which is described by
\begin{align}
\boldsymbol{K}\left(\boldsymbol{X}\right) = \boldsymbol{K}\left(\hat{\boldsymbol{l}}\left(\boldsymbol{X}\right)\right) +\hat{\boldsymbol{\sigma}}_{\mathrm{n}}^2\left(\boldsymbol{X}\right)\,\boldsymbol{I}.
\end{align}
Here, $\hat{\boldsymbol{l}}\left(\boldsymbol{X}\right) \in \mathbb{R}^{N}$ denotes the prediction of the lengthscale parameters for each data point, depending on the PCE coefficients $\boldsymbol{\alpha}_l$ and the input $\boldsymbol{X}$. With $\hat{\boldsymbol{\sigma}}_{\mathrm{n}}^2\left(\boldsymbol{X}\right) \in \mathbb{R}^{N}$ the prediction of the heteroscedastic noise, depending on the PCE coefficients $\boldsymbol{\alpha}_\mathrm{\sigma_\mathrm{n}}$ and the input $\boldsymbol{X}$ is denoted. The task of the optimizer shifts from optimizing the lengthscale  and noise directly to optimizing the polynomial coefficients of the PCE. The gradient from Equation~\eqref{eq_gradient_maximum_likelihood} is now built on the polynomial coefficients for hyperparameter optimization.

\subsection{Runtime Comparison of PCEGP and GP}
In this section, a runtime comparison between the standard GP from GPyTorch~\cite{Gardner.2018} and the PCEGP is carried out. Therefore, three scenarios with $N_\alpha \approx 100$, $N_\alpha=1,000$, and $N_\alpha=10,000$ for the PCEGP are generated, where three different PCEs are used.  For all experiments of the runtime comparison, the squared exponential covariance function is used. The non-stationary lengthscale parameters are calculated with a Legendre polynomial and a uniform distribution in $\left[-1, 1\right]$.

The first configuration is implemented with $n_x=7$ and $q_l=3$, resulting in $N_\alpha~=~120$. To achieve $N_\alpha = 1,001$, a configuration with $n_x=10$ and $q_l=4$ is employed. The third configuration is implemented with $n_x=9$ and $q_l=7$, which results in $N_\alpha = 11,440$. For the standard GP and the DGP, the model training and predictions are performed with $n_x=7$. For all inputs, equidistantly spaced data between $0$ and $1$ are generated. The output is the simple test function $\mathrm{tanh}\left(x\right)$, where $x$ is generated by summarizing each input $x_i$. The experiments are carried out for $N=10$ up to $N=5,000$ with a step size of 50. All experiments are repeated ten times for statistical validation. Each model is trained over 100 gradient descent iterations, with the training time averaged per iteration. The prediction times are calculated by averaging the time required to make predictions for 100 test data points. The PCEGP model is implemented using the base elements of the GPyTorch library for kernel implementations and the SciPy library~\cite{SciPy.2020} for the implementation of the PCE. The standard GP is implemented using the GPyTorch library. The experiments are conducted on a QEMU virtual CPU with 64 virtual CPUs, 32~GB of RAM, and a clock speed of 2.50~GHz. 

\begin{figure}[b!]
    \centering
    \includegraphics[width=0.98\textwidth]{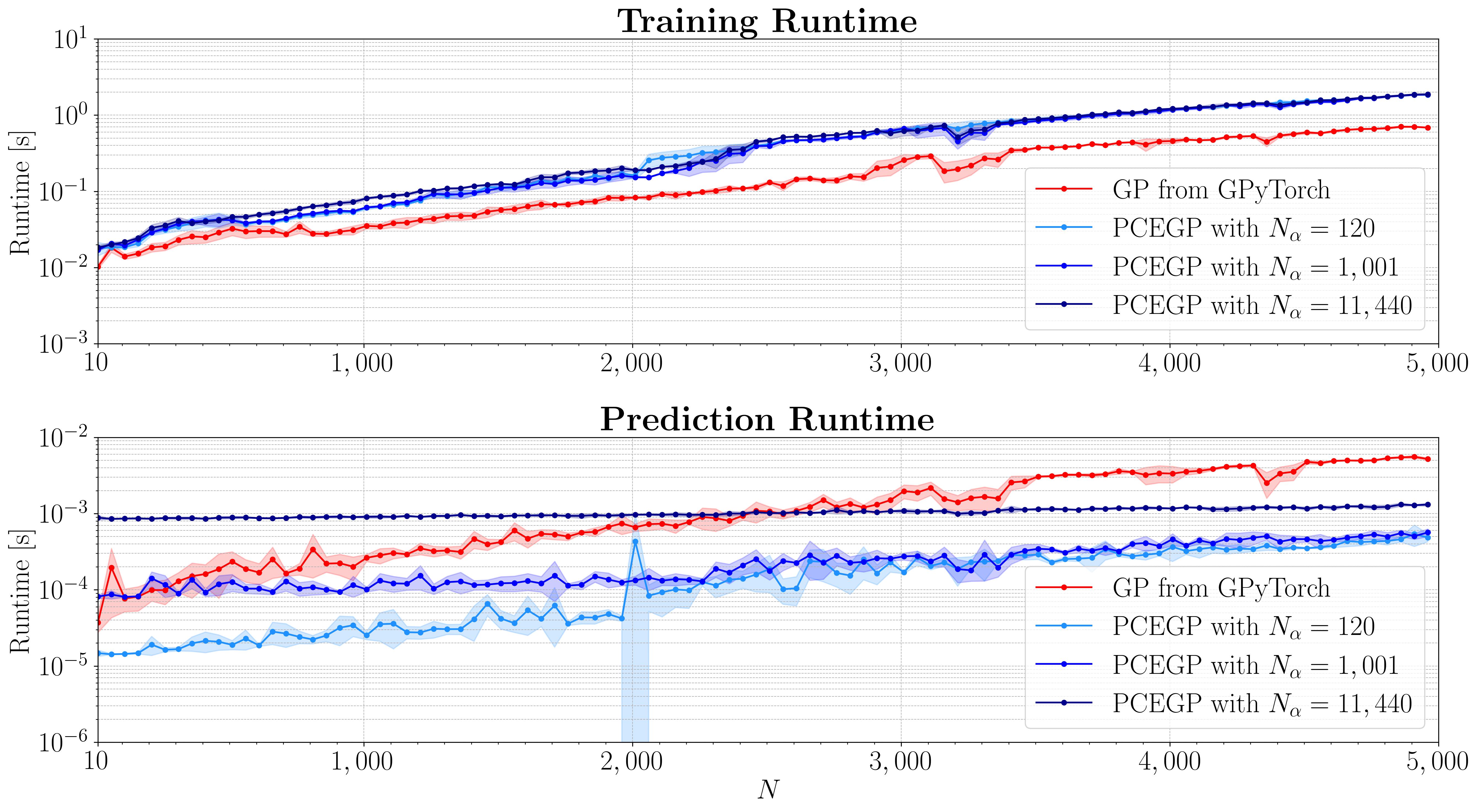}
    \caption{Runtime comparison for training and prediction of PCEGP and GP}
    \label{fig:runtime_comparison}
\end{figure}

The experimental results of the runtime comparison are shown in Figure~\ref{fig:runtime_comparison}. The PCEGP approach exhibits training runtimes slightly above the standard GP from GPyTorch per gradient descent iteration. The faster predictions of the PCEGP model compared to the standard GP from GPyTorch arise from its implementation. The PCEGP model is customized to include only the essential functionalities for mean and variance predictions, avoiding unnecessary computations. 
In contrast, the standard GP model from GPyTorch returns the full multivariate normal distribution for predictions, which, while providing more information, comes with additional computational overhead. Additionally, the standard GP model from GPyTorch includes a broader range of features that are not required in this context, further increasing computational cost. This focused design allows the PCEGP model to remain competitive, even for large parameter spaces like $N_\alpha = 11,440$. 

It can also be observed that the prediction runtime for PCEGP with $N_\alpha = 11,440$ only increases slowly, which indicates a higher overhead of the PCE given the high number of coefficients. However, it should be mentioned that predicting a data point takes only about one millisecond and remains faster than the standard GP from GPyTorch as $N$ increases. For a fairer comparison, the default setting for the maximum number of data points for using the Cholesky decomposition is adjusted to $5,000$ data points.

\section{Experiments for Modeling Non-Stationarity}
This section demonstrates how effectively the PCEGP approach models non-stationary behavior and how its interpretability can be leveraged for deeper insights. This interpretability advantage sets PCEGP apart from nested model structures like the DGP and approaches based on local models such as the HHK GP~\cite{Bitzer.2023}, which do not provide the same level of transparency in capturing locally varying smoothness. To demonstrate the capability of the proposed PCEGP approach to model non-stationarity, a toy example using the Higdon function is used. This function has also been employed by~\cite{Sauer.2023} to evaluate DGPs, highlighting its suitability for testing non-stationary modeling methods. The input $x$ is defined on the interval $x \in [0, 1]$, where the function has two different behaviors for $x < 0.48$ and $x \geq 0.48$. The input data are scaled using the min-max scaler to the range $[-0.5, 0.5]$. This scaling is necessary because the used Legendre polynomials are defined on $[-1, 1]$ and tend to exhibit high function values near the boundaries of this interval. For the outputs, standardization is applied, aligning with the Gaussian process assumption $\boldsymbol{y} \sim \mathcal{N}\left( 0, \boldsymbol{K}\left(\boldsymbol{X}\right) + \sigma^2_{\mathrm{n}},\boldsymbol{I}\right)$. Data scaling is performed using the scikit-learn library~\cite{Pedregosa.2011}. The PCEGP model is implemented using GPyTorch~\cite{Gardner.2018} for kernel computations and SciPy~\cite{SciPy.2020} for PCE. The Nadam optimizer for gradient descent is implemented by Torch~\cite{Novik.2020}. 

The Higdon function is defined as
\begin{equation}
f(x) = 
\begin{cases} 
2 \,\sin(4 \pi \, x) + 0.4 \cos(16 \pi\, x), & \text{for } x < 0.48 \\
2\,x - 1, & \text{for } x \geq 0.48
\end{cases}.
\end{equation}

A polynomial degree of ten with Legendre polynomials is chosen for the PCE component. The PCE values are transformed by the softplus function. The covariance function used for both the PCEGP and the comparison standard GP is the squared exponential covariance function. In Appendix~\ref{pcegp_structure_toy_example}, the model setup for the example experiments is presented.

\begin{figure}[h!]
    \centering
    \includegraphics[width=\textwidth]{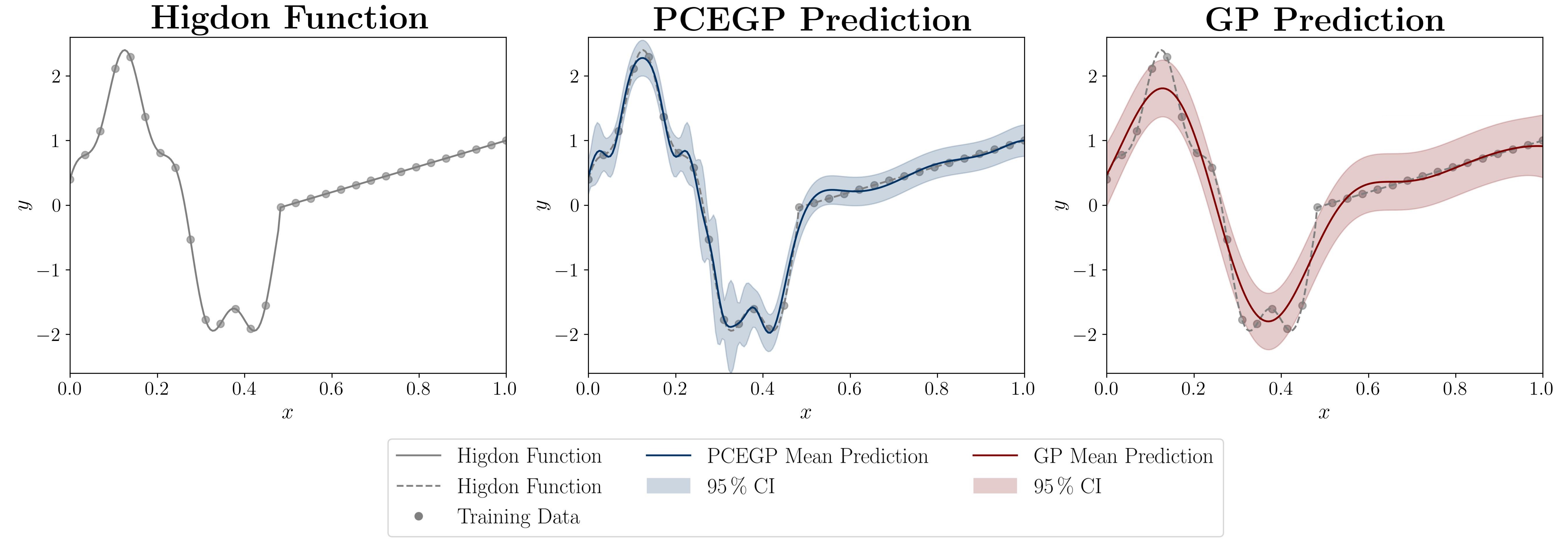}
    \caption{Comparison of PCEGP and GP prediction quality for modeling the non-stationary Higdon function}
    \label{fig:toy_example_higdon1d}
\end{figure}

In Figure~\ref{fig:toy_example_higdon1d}, the prediction results are shown. On the left of Figure~\ref{fig:toy_example_higdon1d}, the original Higdon function is shown with 30 equally spaced training points. To evaluate the model performance, 200 equally spaced test points are used. The middle plot presents the prediction of the PCEGP, which includes the mean prediction and the $95\,\%$ confidence interval (CI) to represent the prediction uncertainty.

The PCEGP effectively captures the non-stationary behavior and locally varying smoothness in the output due to the adaptive lengthscale calculated by the PCE. In contrast, the standard GP prediction, shown on the right, is trained and evaluated on the same data. The GP exhibits stationary smoothness and fails to adapt to the local changes in the Higdon function.

In Table~\ref{tab:error_metrics_higdon1d}, the error metrics used to evaluate the model performance on the test data are presented (detailed explanations of these metrics are provided in Appendix~\ref{error_metrics}). The PCEGP approach shows better performance in capturing the non-stationary behavior compared to the GP, as reflected in the metrics.

\begin{table}[h!]
    \centering
    \caption{Error metrics for PCEGP and GP model performance on test data of Higdon function}
    \begin{tabular}{lccccc}
        \toprule
        \textbf{Model} & \textbf{MAE} & \textbf{MedAE} & \textbf{MSE} & \textbf{RMSE} & \textbf{R$^2$} \\
        \midrule
        PCEGP & 0.059 & 0.044 & 0.006 & 0.079 & 0.995 \\
        GP    & 0.161 & 0.095 & 0.053 & 0.231 & 0.953 \\
    \end{tabular}
    \label{tab:error_metrics_higdon1d}
\end{table}

To further demonstrate the ability of the proposed PCEGP approach to model non-stationary behavior and provide insights into input sensitivity, the Higdon function is extended to two dimensions. The first dimension, $x_1$, retains the original Higdon function, while a simple linear function is introduced for the second dimension, $x_2$, resulting in $f(x_1, x_2)$ as

\begin{equation}
f(x_1, x_2) = 
\begin{cases} 
2 \,\sin(4 \pi \,x_1) + 0.4 \,\cos(16 \pi x_1) + x_2, & \text{for } x_1 < 0.48 \\
2\, x_1 - 1 + x_2, & \text{for } x_1 \geq 0.48 
\end{cases}.
\end{equation}

The function is defined over the interval $x_1, x_2 \in [0, 1]$. In this experiment, 200 training data points are randomly sampled from a uniform distribution over this domain, and $1,000$ test points are used to evaluate the model performance. The configuration of the PCEGP and GP model are the same as in the first experiment. The second example emphasizes the interpretability advantage of the PCEGP in capturing non-stationary behavior in the output due to varying lengthscale parameters. Such interpretability is not possible with nested model structures like the DGP or with approaches consisting of different local models like the HHK GP~\cite{Bitzer.2023}. This makes the PCEGP approach particularly useful for understanding locally varying smoothness in the model output.

\begin{figure}[b!]
    \centering
    \includegraphics[width=\textwidth]{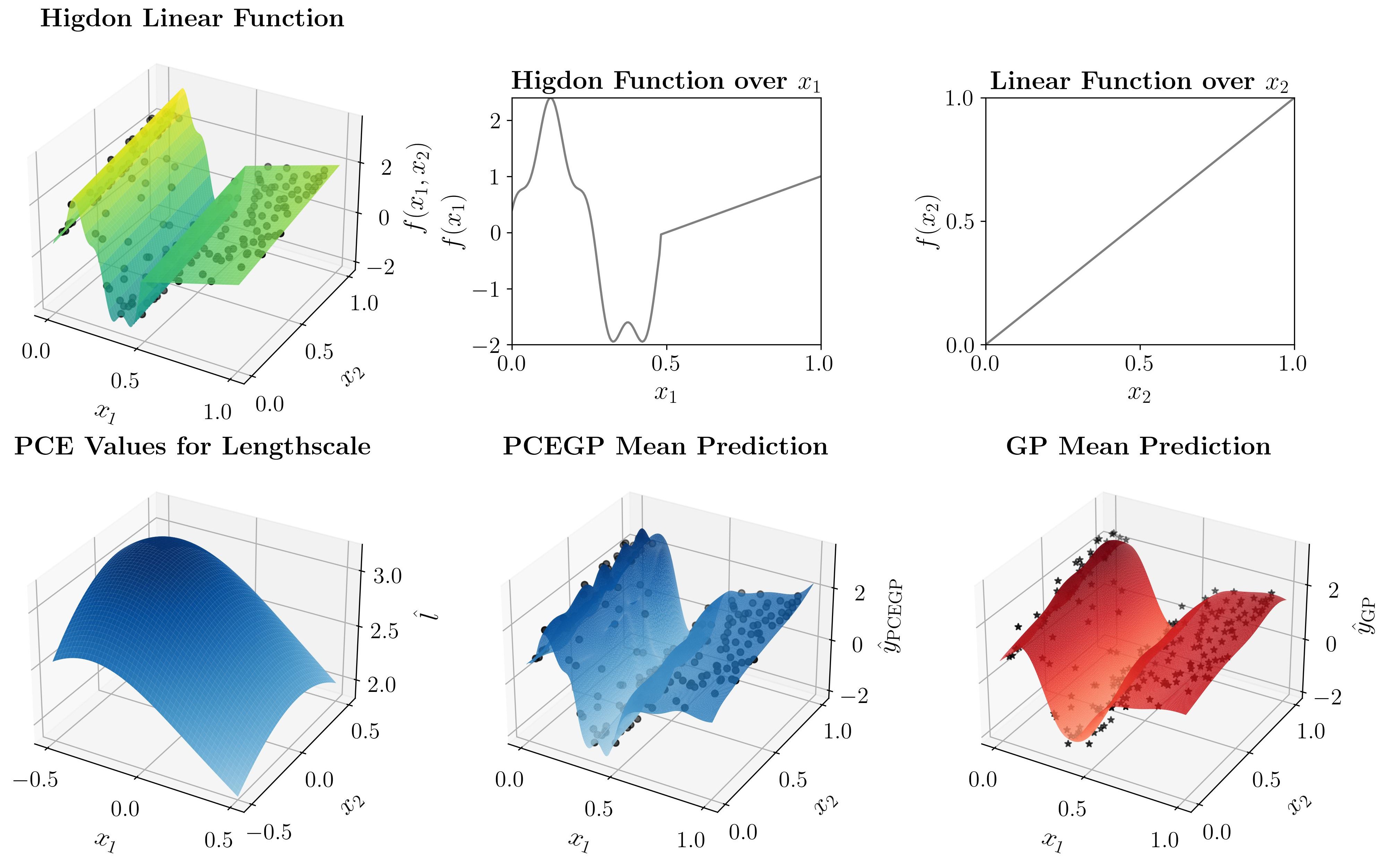}
    \caption{Comparison of PCEGP and GP prediction quality for modeling the non-stationary Higdon function}
    \label{fig:toy_example_higdon2d}
\end{figure}

In Figure~\ref{fig:toy_example_higdon2d}, the results are shown. The extended Higdon function with training data is shown in the top left plot. The top middle plot shows the function along $x_1$ with $x_2=0$, and the top right plot shows the function along $x_2$ with $x_1=0.5$. The lower right plot shows the GP mean prediction, which struggles to represent the non-stationary behavior in $x_1$. The PCEGP mean prediction is shown in the lower middle plot and the PCE calculated lengthscale parameters are shown in the lower left plot, over the scaled input range~$[-0.5, 0.5]$.
The lengthscale values calculated by the PCE indicate stronger fluctuations along~$x_1$, suggesting that~$x_1$ is the input responsible for the non-stationary behavior, while~$x_2$ exhibits stationary behavior. This observation can be quantified by calculating the variance of the lengthscale parameter across the input space for~$x_1$ and~$x_2$, normalized by the total variance over the entire domain. Specifically, for each fixed value of~$x_2$, the variance along the~$x_1$-axis is determined, and similarly, for each fixed value of~$x_1$, the variance along the $x_2$-axis is calculated. This can be expressed as
\begin{align}
\operatorname{Var}_{x_1}(\hat{l}\left(\boldsymbol{x}\right)) &= \frac{1}{M} \sum_{j=1}^{M} \left( \hat{l}\left( x_{1}, x_{2,j}\right) - \frac{1}{M} \sum_{k=1}^{M} \hat{l}\left(x_{1}, x_{2,k}\right) \right)^2, \\
\operatorname{Var}_{x_2}(\hat{l}\left(\boldsymbol{x}\right)) &= \frac{1}{M} \sum_{i=1}^{M} \left( \hat{l}\left(x_{1,i}, x_{2} \right) - \frac{1}{M} \sum_{k=1}^{M} \hat{l}\left(x_{1,k}, x_{2}\right) \right)^2,
\end{align}
where $M \in \mathbb{N}$ denotes the number of points evaluated across each axis. Here, $\operatorname{Var}_{x_1}\left(\hat{l}\left(\boldsymbol{x}\right)\right)$ represents the average variance over the values of $x_1$, evaluated for each fixed $x_{2,j}$, and similarly, $\operatorname{Var}_{x_2}\left(\hat{l}\left(\boldsymbol{x}\right)\right)$ reflects the average variance over $x_2$, evaluated for each fixed $x_{1,i}$. 

These variance components are then normalized to determine the relative importance of each input in driving the non-stationary behavior
\begin{align}
\text{Total variance: } & \quad \mathrm{Var}_{\text{total}} = \operatorname{Var}_{x_1}\left(\hat{l}\left(\boldsymbol{x}\right)\right) + \operatorname{Var}_{x_2}\left(\hat{l}\left(\boldsymbol{x}\right)\right)\hspace{-1mm}, \\
\text{Normalized variance for } x_1: & \quad \mathrm{Var}_{x_1, \mathrm{N}} = \frac{\operatorname{Var}_{x_1}\left(\hat{l}\left(\boldsymbol{x}\right)\right)}{\mathrm{Var}_{\text{total}}} = 0.922, \\
\text{Normalized variance for } x_2: & \quad \mathrm{Var}_{x_2,\mathrm{N}} = \frac{\operatorname{Var}_{x_2}\left(\hat{l}\left(\boldsymbol{x}\right)\right)}{\mathrm{Var}_{\text{total}}} = 0.078.
\end{align}

These metrics $\mathrm{Var}_{x_1, \mathrm{N}}$ and $\mathrm{Var}_{x_2, \mathrm{N}}$ represent the normalized variances for $x_1$ and $x_2$, indicating the relative contributions of each input to the non-stationarity in the output. The analysis shows that $x_1$ contributes more significantly to the non-stationary behavior due to its higher variance, as calculated from the lengthscale parameters.

In Table~\ref{tab:error_metrics_higdon2d}, the error metrics used to assess the model performance on the test data are presented. The PCEGP approach outperforms the GP across all metrics, confirming its ability to model non-stationarity.

\begin{table}[h]
    \centering
    \caption{Error metrics for PCEGP and GP model performance on test data of the extended Higdon function}
    \begin{tabular}{lccccc}
        \toprule
        \textbf{Model} & \textbf{MAE} & \textbf{MedAE} & \textbf{MSE} & \textbf{RMSE} & \textbf{R$^2$} \\
        \midrule
        PCEGP & 0.110 & 0.047 & 0.037 & 0.192 & 0.970 \\
        GP    & 0.194 & 0.120 & 0.077 & 0.277 & 0.937  \\
    \end{tabular}
    \label{tab:error_metrics_higdon2d}
\end{table}

\section{Benchmark Experiments\label{experiments}}
In this section, the presented approach is applied to six different regression benchmark datasets from the UCI machine learning repository\footnote{\url{https://archive.ics.uci.edu/ml/index.php}}. The benchmark regression datasets are explained in Appendix~\ref{uci_datasets}. The PCEGP model performance is evaluated against three stationary and two non-stationary GP-based approaches, which are presented in the following. All data are scaled in the same way for model training.  The inputs are scaled to a standard range of $[-1, 1]$ using a min-max scaler for all compared methods, except for the PCEGP model, where the inputs are scaled to the range $[-0.5, 0.5]$. This adjustment is necessary because Legendre polynomials are defined on $[-1, 1]$ and exhibit high function values near the boundaries of this range.  For the outputs, standardization of data is applied, as this corresponds to the assumption $\boldsymbol{y} \sim \mathcal{N}\left( 0, \boldsymbol{K}\left(\boldsymbol{X}\right) + \sigma^2_{\mathrm{n}}\,\boldsymbol{I}\right)$ for GPs. For data scaling the scikit-learn library~\cite{Pedregosa.2011} is used. 

The experimental results are generated using a ten-fold cross-validation. All datasets are segmented into ten equally sized, non-overlapping subsamples through random selection. One of these subsamples is assigned as validation data, while the remaining nine subsamples are utilized as training data. This process is iterated ten times, with each subsample taking on the role of the validation set once. For statistical validation, the experiments are repeated ten times. A different random seed is selected in each repetition. The aim is to achieve the smallest possible mean prediction error across all repetitions. The experiments are run on a QEMU virtual CPU with 64 virtual CPUs and 32~GB~RAM at 2.50~GHz. 

\subsection{PCEGP Model Setup}
The PCEGP model is implemented using the GPyTorch library~\cite{Gardner.2018} for kernel implementations and the SciPy library~\cite{SciPy.2020} for the implementation of the PCE. The TPE is implemented, using the Optuna library~\cite{Akiba.2019} and the gradient descent optimizer Nadam is implemented using the Torch library~\cite{Novik.2020}. The Nadam optimizer is preferred over the Adam optimizer because it achieves faster convergence through the Nesterov momentum, as described in \cite{dozat2016nadam}, making it particularly suitable given the large number of PCE coefficients.  

All experiments are carried out with the same model structure, which is illustrated in Appendix~\ref{pcegp_structure_experiments}. The lengthscale parameters $\hat{\boldsymbol{l}}_{i}\left(\boldsymbol{x}\right)$ of each covariance function are calculated by the Legendre polynomials with a uniform distribution between $-1$ and $1$ and a fixed polynomial degree $q_l =5$. For the experiments, no hyperparameter transformation is used for the lengthscale as this turned out, to be the best setup for the UCI datasets. Hence, the methodical investigation of the hyperparameter transformation is not part of this work. Here, the covariance structure is build of the squared exponential, absolute exponential, Matérn$_{3/2}$, Matérn$_{5/2}$, and rational quadratic covariance function for the GP. This covariance structure has shown good results in~\cite{Cremanns.2017}.  The noise variance $\hat{\sigma}_\mathrm{n}^2\left(\boldsymbol{x}\right)$ is homoscedastic. 

In this work, the TPE is used to suggest the PCE coefficients $\alpha_0$ for each covariance function before the gradient descent optimization, which represents the mean value of the non-stationary lengthscale. This minimizes the risk of getting trapped in local minima. In addition, the learning rate $\eta$ and the number of gradient descent iterations $n_\mathrm{I}$, which are crucial for a highly parametric model are suggested by the TPE. For all datasets, the HPO is performed with random search of 100 HP combinations, followed by the use of the TPE for further 100 HP combinations with a ten-fold cross validation for the gradient descent optimization. The evaluation criterion for the Bayesian HPO is the mean root mean squared error (RMSE) of the ten validation datasets. The best combination of $\alpha_0$, $\eta$, and $n_\mathrm{I}$ is used for all repetitions in the followed ten-fold cross validation. In Table~\ref{tab:parameter_bounds}, the parameter bounds for HPO are shown. These bounds are used for all datasets, except the wine quality dataset. For the wine quality dataset, the maximum number of gradient iterations is limited to 500, as training often failed for higher values due to a non-positive semidefinite covariance matrix.

\begin{table}[h!]
\centering
\caption{Hyperparameter bounds for Bayesian HPO}
\begin{tabular}{ll}
\hline
\textbf{Parameter}  & \textbf{Values} \\ \hline
Learning rate $\eta$ & 0.001, 0.0025, 0.005, 0.01, 0.05, 0.1 \\ 
Gradient descent iterations $n_\mathrm{I}$ & 50, 100, 200, 300, 400, 500, 600, 700, 800, 900, $1000$\\
PCE coefficient $\alpha_0$ & 0.1 to 2.0 (step size: 0.1) \\ \hline
\end{tabular}
\label{tab:parameter_bounds}
\end{table}

\subsection{Compared Methods}
In the following, the compared methods of this work are presented. All used methods are implemented with homoscedastic noise.

\textbf{Gaussian Process (GP)}~\cite{Rasmussen.2006}: This method employs a stationary model with a squared exponential covariance function and utilizes the Adam optimizer from Torch~\cite{Novik.2020} for gradient descent optimization with $\eta = 0.1$ and $n_\mathrm{I} = 100$. The model is implemented using the GPyTorch library~\cite{Gardner.2018}.

\textbf{Gaussian Process with ARD (GP-ARD)}~\cite{Rasmussen.2006}: This stationary GP model also employs a squared exponential covariance function, with a separate lengthscale parameter for each input to adjust for input relevance. The Adam optimizer from Torch~\cite{Novik.2020} is employed for gradient descent optimization with $\eta = 0.1$ and $n_\mathrm{I} = 100$. This model is also implemented using the GPyTorch library~\cite{Gardner.2018}.

\textbf{Deep Gaussian Process (DGP)}~\cite{Damianou.2013}: The DGP is a non-stationary model that applies a squared exponential covariance function to each of its two layers, using 50 inducing points. For each layer the number of nodes matches to the number of inputs in the training data. As demonstrated in~\cite{Sauer.2023}, this configuration, consisting of two layers with a number of nodes matching the input dimension, performed well in~\cite{Sauer.2023}. The Adam optimizer from Torch~\cite{Novik.2020} is utilized for gradient descent optimization with $\eta = 0.001$ and $n_\mathrm{I} = 5,000$. In this work, mini-batch training is not used for this approach, as it has not shown any advantages in preliminary tests and is significantly more time-consuming.  The DGP is implemented using the GPyTorch library~\cite{Gardner.2018}.

\textbf{Hierarchical Hyperplane Kernel for GPs}~\cite{Bitzer.2023}: This method is non-stationary and uses hyperplanes to partition the input space, with latent GPs applied to each partition. Each latent GP has a squared exponential covariance function, with four latent GPs balancing complexity and performance. In the case that the approach with four latent GPs does not converge, two latent GPs are used. The L-BFGS optimizer~\cite{Bottou.2018} from TensorFlow~\cite{TensorFlowDevelopers.2024} is used for gradient descent optimization, and the method is implemented using the Python-based HHK library\footnote{\url{https://github.com/boschresearch/Hierarchical-Hyperplane-Kernels}}.

\textbf{GPBoost}~\cite{Sigrist.2022}: The GPBoost approach combines tree boosting and a stationary GP model using an absolute exponential covariance function. A combination of grid search and gradient descent is utilized for hyperparameter optimization, with grid search applied to parameters such as the tree structure, number of boosting iterations, learning rate, minimum number of data per leaf, and maximum tree depth, while gradient descent is used for further optimization of the GPs parameters. The HPO is carried out in an inner loop four-fold cross-validation for each fold of the outer loop ten-fold cross-validation with the same HP grid as shown in~\cite{Sigrist.2022}. In~\cite{Sigrist.2022}, the GPBoost model is optimized differently for spatial datasets such as Boston housing. An out-of-sample evaluation is performed there for the HPO. In this work, the GPBoost model is trained on all datasets with the same HPO, which means that the results do not correspond to those of~\cite{Sigrist.2022}. The method is implemented using the Python package of the GPBoost library\footnote{\url{https://github.com/fabsig/GPBoost}}.

\subsection{Results and Discussion}
\begin{figure}[t!]
    \centering
    \includegraphics[width=\textwidth]{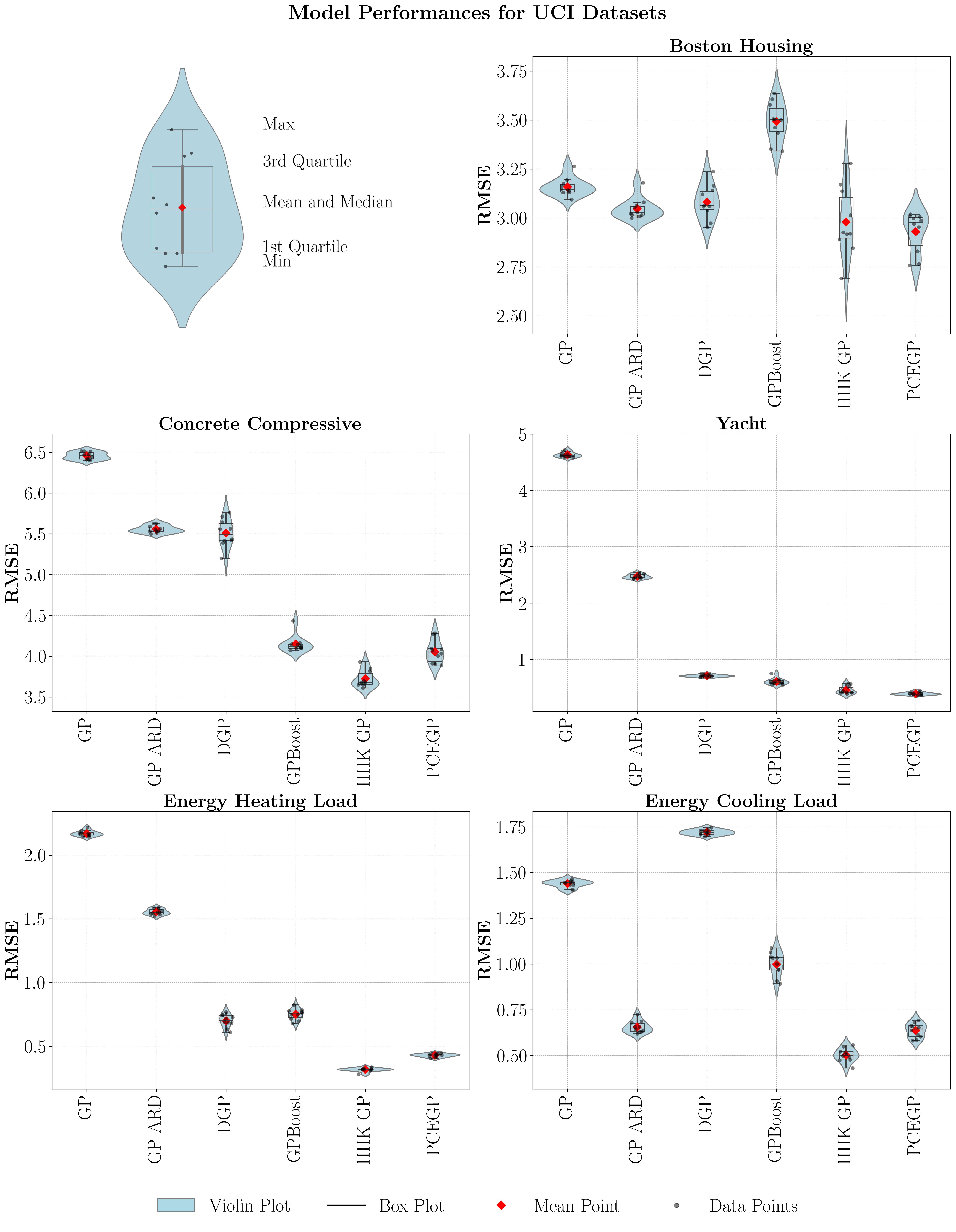}
    \caption{RMSE for UCI Datasets of the Compared Methods and PCEGP}
    \label{fig:result_plot_rmse1}
\end{figure}

\begin{figure}[t!]
    \centering
    \includegraphics[width=\textwidth]{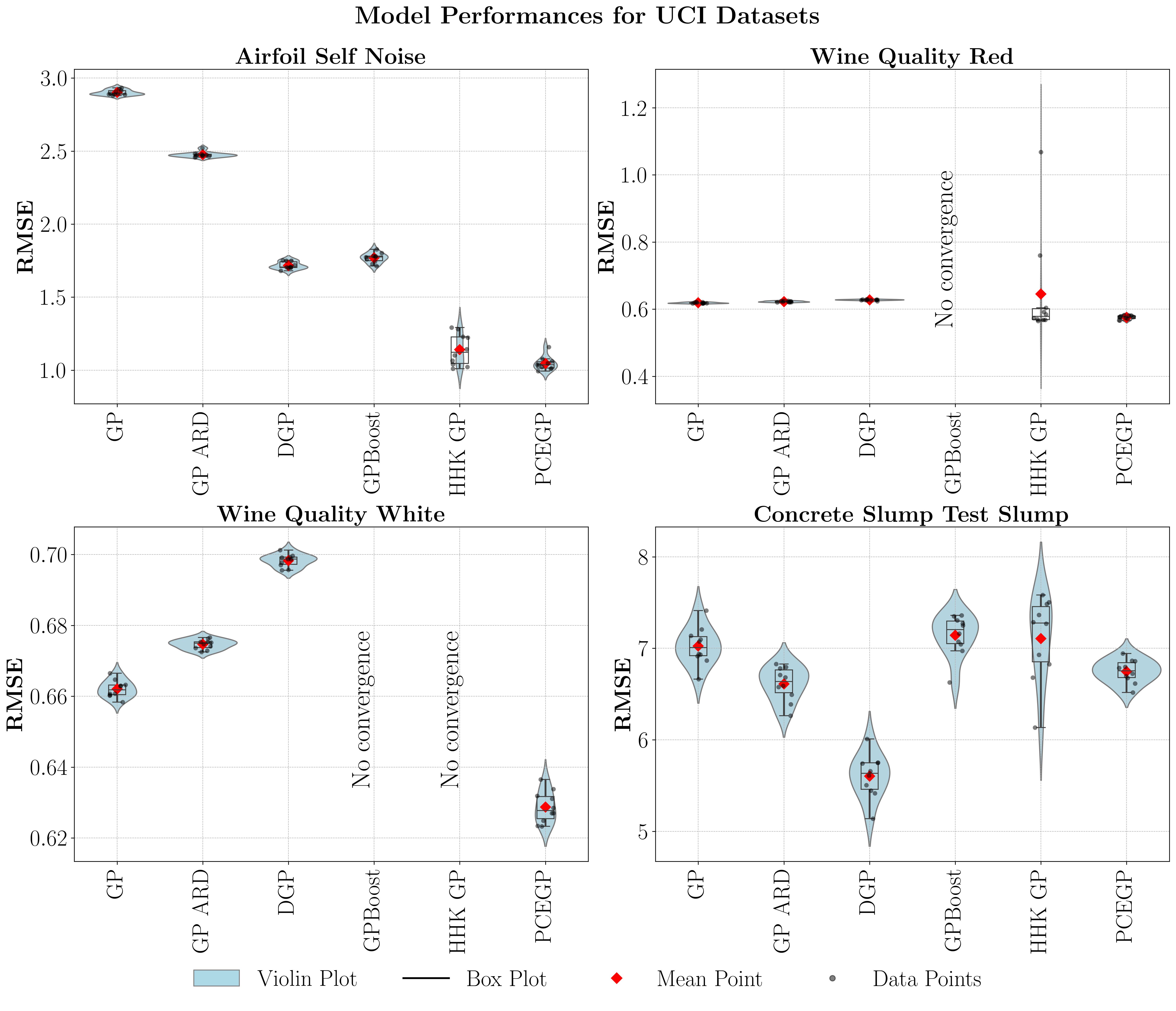}
    \caption{RMSE for UCI Datasets of the Compared Methods and PCEGP}
    \label{fig:result_plot_rmse2}
\end{figure}

In this section, the results of the model performances on the benchmark tests are presented. The model performances are evaluated using five different error metrics, which are explained in detail in Appendix~\ref{error_metrics}. The following discussion focuses specifically on the root mean squared error (RMSE) shown in Figure~\ref{fig:result_plot_rmse1} and Figure~\ref{fig:result_plot_rmse2}. A comprehensive overview of model performances for all error metrics across the UCI datasets is provided in Appendix~\ref{appendix_results}.

In Table~\ref{tab:datasets}, an overview of the datasets used in this work is given, summarizing the number of data points $N$, input variables $n_x$, and outputs for each dataset. These datasets cover a wide range of data sizes and input dimensions, allowing for a comprehensive evaluation of the GP-based models. 

\begin{table}[h!]
\centering
\caption{Overview of the datasets}
\begin{tabular}{@{}lcccccccc@{}}
\toprule
\textbf{Dataset} & \textbf{Boston} & \textbf{Concrete} & \textbf{Energy} & \textbf{Airfoil} & \textbf{Yacht} & \textbf{Wine} & \textbf{Slump}\\ 
\midrule
$N$ & $506$ & $1,030$ & $768$ & $1,503$ & $308$ & $6,497$ & $103$ \\ 
$n_x$& 13  & 8  & 8   & 5     & 6   & 11 & 7\\
Outputs & 1 & 1 & 2 &  1 & 1 & 2 & 3 \\ 
\bottomrule
\end{tabular}
\label{tab:datasets}
\end{table}

In the upper left of Figure~\ref{fig:result_plot_rmse1}, the explanation of the result plots is illustrated. Each violin plot is combined with a box plot to provide a detailed representation of the RMSE distribution. Individual points indicate the mean RMSE values from each ten-fold cross-validation repetition, while a distinct marker highlights the overall mean across all repetitions. The horizontal line within each box plot denotes the median RMSE, with the first and third quartiles, as well as the minimum and maximum values, also displayed. The violin plot itself visualizes the overall distribution of RMSE values, offering insights into variability and spread in model performance.

In Appendix~\ref{appendix_results}, a comprehensive tabular overview of the model performances for six commonly used error metrics is given. In Figure~\ref{fig:result_plot_rmse1} and Figure~\ref{fig:result_plot_rmse2}, the performances of the GP-based models on benchmark datasets are shown, considering the RMSE. Analyzing the RMSE results, the proposed PCEGP approach outperforms all other approaches in five out of nine benchmarks.

On all other datasets, PCEGP performs comparably to the non-stationary HHK GP approach which is the second best approach in the comparison. It should be noted that the HHK GP approach on the Airfoil Self Noise and Wine Quality Red datasets does not converge with four latent GPs and in these two cases the model structure is set to two latent GPs to achieve convergence. It should also be noted that the HHK GP approach on the Wine Quality White dataset did not converge with either eight, four, or two latent GPs. Similarly, the GPBoost approach failed to converge on the Wine Quality Red and Wine Quality White dataset. The PCEGP approach, on the other hand, achieves convergence and good results on all data sets with the same model structure. Another advantage of the PCEGP approach is that its repeatability is in most cases better than that of the HHK GP approach, as can be seen from the distributions in the result plots.

The PCEGP consistently outperforms the standard GP, GP with ARD, and the state-of-the-art method GPBoost. The non-stationary DGP is outperformed on eight of nine benchmarks. These results demonstrate the robustness and adaptability of PCEGP across diverse real-world datasets, including small and large datasets as well as low- and high-dimensional feature spaces. This highlights its potential as an effective tool for modeling complex systems. One aspect that needs to be improved is the uncertainty estimation of the PCEGP model, as the Test NLL metric does not yield consistent good results across all datasets, and outliers sometimes appear.

\clearpage

\section{Summary and Outlook\label{summary_and_outlook}}
This paper presents a novel ML approach for modeling regression tasks. The Polynomial Chaos Expanded Gaussian Process is introduced to model the global and local behavior of data in one model through non-stationarity. This is achieved by calculating the lengthscale parameters of the GP's covariance functions as a function of input data using PCE. In addition, heteroscedastic noise estimation based on PCE is introduced.

The novel approach proved to be competitive on various benchmark datasets for regression and outperforms previous stationary and non-stationary GP-based approaches. The hyperparameters obtained from PCE are expressed analytically as polynomials. This characteristic enhances model interpretability, especially when compared to other non-stationary GP-based methods such as DGP and HHK~GP. Furthermore, by analyzing the variation in the functional values of the PCE, it is possible to identify which input variables are responsible for inducing non-stationary behavior in the output. This capability highlights an additional strength of the proposed method, providing valuable insights into input-output relationships in complex systems.

The algorithm currently also has limitations, as expert knowledge is required to search for the optimal parameters in order to restrict the search space of these parameters. It is also necessary to manually select a suitable data scaler, polynomial basis, and covariance function for the data. Additionally, as the number of inputs increases, the number of coefficients also grows, which can cause challenges in the optimization process.

Further research is needed on the automation of the selection of the data scaler, polynomial bases, and covariance functions. Another point for future research is the investigation of the transformations for the lengthscale and noise variance and what influence these have on the model performance. The suitability of PCEs for small datasets makes an investigation and comparison of model performance on adaptive learning tasks another point for future work. In addition, future work will investigate the performance of the PCEGP for time series forecasting and modeling of dynamic systems.

\bibliographystyle{ieeetr}

\bibliography{literature}
\clearpage

\begin{appendices}

\section{Notations}
The notations used throughout this paper are summarized in Table~\ref{notation_table}. This table serves as a reference for the mathematical symbols and terms.
\begin{longtable}{p{0.22\textwidth}|p{0.73\textwidth}}
	
  \caption{Notations}\label{notation_table} \\
  \textbf{Formula symbol} & \textbf{Description} \\ \hline\hline
  \endhead
  \endfoot
  \endlastfoot
  \vspace{-2mm}$a(h | \mathcal{D}_{\mathrm{HP}})$ & \vspace{-2mm}Acquisition function derived from the ratio of KDEs in $\mathbb{R}$ \\ 
  $b^{(b)}, b^{(w)}$ & Bandwidth parameters for the KDEs of the better \\
                     & and worse groups in $\mathbb{R}^+$ \\ 
  $b$ & Number of polynomial bases\\
  EI & Expected improvement acquisition function\\ 
    $f(\boldsymbol{\xi})$ & Function approximated by the PCE, depending \\
                          & on the random input variables $\boldsymbol{\xi}$ \\
  $f\left(x^*\right)$ & Observed minima in $\mathbb{R}$\\
  $\boldsymbol{g}_i$ & Gradient of the loss function $\mathcal{L}$ at iteration $i$ in $\mathbb{R}^{N_{\mathrm{\theta}}}$ \\  
  $\boldsymbol{h}_i$ & Hyperparameter configuration at the $i$-th observation in $\mathbb{R}^{N_{\mathrm{HP}}}$ \\
  $\boldsymbol{h}$ & Set of hyperparameters in $\mathbb{R}^{N_{\mathrm{HP}}}$ \\
  $\boldsymbol{h}_{\mathrm{opt}}$ & Optimal set of hyperparameters that minimizes \\
                                  & the objective function in $\mathbb{R}^{N_{\mathrm{HP}}}$ \\
  $\boldsymbol{I}$ & Identity matrix \\
  $k_i$ & Degree of the univariate orthogonal polynomial $\phi_{k_i}(\xi_i)$ associated with the random variable $\xi_i$ in   
$\mathbb{N}_0$ \\  
  $\boldsymbol{k}$  & Multi-index vector, $\boldsymbol{k} = (k_1, k_2, \ldots, k_{n_\xi})$, representing the degrees of the polynomial basis functions \\
                    & corresponding to each random variable in $\mathbb{N}^{n_\xi}$ \\
  $k_{\mathrm{KDE}}$ & Kernel function for density estimation\\
  $k(\boldsymbol{x},\boldsymbol{x}')$ & Covariance function of a Gaussian process between $\boldsymbol{x}$ and $\boldsymbol{x}'$\\
  $k_\mathrm{\Sigma}(\boldsymbol{x},\boldsymbol{x}')$ & Summation of $n_k$ covariance functions\\  
  
  $\boldsymbol{k}\left(\boldsymbol{X}, \boldsymbol{x}^*\right)$ & Covariance vector depending on the training data \\
                                                                & and the next point to predict in $\mathbb{R}^{N \times 1}$\\  
  $\boldsymbol{K}\left(\boldsymbol{X}\right)$ & Covariance matrix depending on the training data in $\mathbb{R}^{N \times N}$\\
  $l$ & Lengthscale parameter of a GP in $\mathbb{R}^{+}$\\
  $l\left(x\right)$ & Data point dependent lengthscale parameter of a GP in $\mathbb{R}$\\
  $l_i$ & Lengthscale parameter of a GP at input $i$ in $\mathbb{R}^{+}$\\  
  $\hat{l}\left(x\right)$ & Prediction of the lengthscale parameter by the PCE in $\mathbb{R}$\\  
  $\hat{l}_\mathcal{T}\left(x\right)$ & Prediction of the transformed lengthscale parameter \\
  									 & by the PCE in $\mathbb{R}$\\  
  $\hat{\boldsymbol{l}}\left(\boldsymbol{x}\right)$ & Prediction of the lengthscale parameter vector by the PCE in $\mathbb{R}^{N}$ \\
  $\boldsymbol{L}$ & Lengthscale parameter diagonal matrix of a GP \\ 
                   & which includes all $l_i$ lengthscale parameters in $\mathbb{R}^{n_x \times n_x}$ \\  
  L1 & Lasso penalty\\
  L2 & Ridge penalty\\
  \vspace{-2mm}$m(\boldsymbol{x})$ & \vspace{-2mm}Mean function of a GP for input vector $\boldsymbol{x}$\\
  $m_i$ &  First moment estimate of the gradient \\
        & at iteration $i$ in $\mathbb{R}^{N_{\mathrm{\theta}}}$ \\  
  $\hat{m}_i$ & Bias-corrected first moment estimate \\
              & at iteration $i$ in $\mathbb{R}^{N_{\mathrm{\theta}}}$ \\
  $M$ & Number of test points in $\mathbb{N}$\\ 
  $n_\mathrm{F}$ & Number of folds in $\mathbb{N}$\\   
  $n_\mathrm{I}$ & Number of gradient descent iterations in $\mathbb{N}$ \\
  $n_\mathrm{IT}$ & Number of initial trials in $\mathbb{N}$\\
  $n_k$ & Number of covariance functions in $\mathbb{N}$\\
  $n_\mathrm{T}$ & Number of trials in $\mathbb{N}$\\     
  $n_x$ & Number of input parameters in $\mathbb{N}$\\
  $n_\xi$  & Number of random input variables in the PCE in $\mathbb{N}$\\ 
  $N$ & Number of training points in $\mathbb{N}$\\
  $N_{\alpha}$  & Number of coefficients in the truncated PCE in $\mathbb{N}$\\
  $N_{\alpha,l}$ & Number of coefficients for \\
                 & lengthscale estimation in $\mathbb{N}$\\
  $N_{\alpha,\sigma_{\mathrm{n}}}$ & Number of coefficients for noise estimation in $\mathbb{N}$\\
  $N_{\mathrm{obs}}$ & Number of observations in $\mathbb{N}$ \\ 
  $N_{\mathrm{HP}}$ & Number of hyperparameters in $\mathbb{N}$ \\ 
  $N_{\mathrm{layer}}$ & Number of layers for DGP in $\mathbb{N}$ \\ 
  $N_{\mathrm{\theta}}$ & Number of model hyperparameters in $\mathbb{N}$ \\
  $o_{\mathrm{B},i}$ & Objective function value of the $i$-th observation in $\mathbb{R}$ \\ 
  $o_\mathrm{B}(\boldsymbol{h})$ & Objective function representing the performance \\
                                 & of a ML model based on hyperparameters in $\mathbb{R}$ \\
  $p\left(\xi\right)$ & Probability density function associated with the \\
                      & random input variable $\xi$\\
  $p(\boldsymbol{\xi})$ & Joint probability density function associated \\
                        & with the random input variables $\boldsymbol{\xi}$ \\
  $p(h|\mathcal{D}_{\mathrm{HP}}(b))$ & Probability distribution for the "better group"  \\
  $p(h|\mathcal{D}_{\mathrm{HP}}(w))$ & Probability distribution for the "worse group" \\
  $q$  & Maximum polynomial degree in the truncated PCE in $\mathbb{N}_0$ \\
  $q_l$ & Polynomial degree of the lengthscale parameter estimation\\
  $q_{\sigma_{\mathrm{n}}}$ & Polynomial degree of the noise variance estimation\\
  $v_i$ & Second moment (variance) estimate of the gradient\\
  & at iteration $i$ in $\mathbb{R}^{N_{\mathrm{\theta}}}$ \\  
  $\hat{v}_i$ & Bias-corrected second moment estimate \\
  			& at iteration $i$ in $\mathbb{R}^{N_{\mathrm{\theta}}}$ \\
  $\boldsymbol{v}$ & Trivial vector in $\mathbb{R}^{n_x}$\\
  
$\operatorname{Var}_{x_i}(\hat{l}(\boldsymbol{x}))$ & Variance of the lengthscale parameter $\hat{l}(\boldsymbol{x})$ \\
													&along the $x_i$-axis in $\mathbb{R}^{+}$\\
\vspace{-2mm}$\operatorname{Var}_{\text{total}}$ & \vspace{-2mm}Total variance, computed as the sum of variances \\
&along all input dimensions \\
$\operatorname{Var}_{x_i, \mathrm{N}}$ & Normalized variance along the $x_i$-axis, indicating \\
& its relative contribution to the total variance \\

  $x$ & Scalar training input  in $\mathbb{R}$\\
  $x^*$ & Scalar test input in $\mathbb{R}$\\
  $x_\mathrm{s}$ & Scaled scalar training input in $\mathbb{R}$\\
  $x_\mathrm{s}^*$ & Scaled scalar test input  in $\mathbb{R}$\\
  $\boldsymbol{x}$ & Training input vector in $\mathbb{R}^{n_x}$ \\
  $\boldsymbol{x}^*$ &  Test input vector in $\mathbb{R}^{n_x}$ \\
  $\boldsymbol{x}_\mathrm{s}$ & Scaled training input vector in $\mathbb{R}^{n_x}$ \\
  $\boldsymbol{x}_\mathrm{s}$ & Scaled test input vector in $\mathbb{R}^{n_x}$ \\
  $\boldsymbol{x}_i$ & $i$-th Training input vector in $\mathbb{R}^{n_x}$ \\ 
  $\boldsymbol{X}$ & Training input matrix in $\mathbb{R}^{N \times n_x}$\\
  $y$ & Training output value in $\mathbb{R}$ \\
  $y_\mathrm{s}$ & Scaled training output value in $\mathbb{R}$ \\
  $\boldsymbol{y}$ & Training output vector in $\mathbb{R}^{N}$\\
  $\boldsymbol{y}_\mathrm{s}$ & Scaled training output vector in $\mathbb{R}^{N}$\\
  $y_i$ & $i$-th training output value in $\mathbb{R}$ \\
  $\hat{\boldsymbol{y}}_\mathrm{PCEGP}$ & PCEGP model mean and uncertainty prediction \\
                                        & in $\mathbb{R}^{2 \times 1}$\\
  $\hat{\boldsymbol{y}}_\mathrm{PCEGP s}$ & Scaled PCEGP model prediction in $\mathbb{R}^{2\times 1}$\\
  $\hat{y}_\mathrm{PCEGP s,m}$ & Scaled PCEGP model mean prediction in $\mathbb{R}$\\
  $\hat{y}_\mathrm{PCEGP s,v}$ & Scaled PCEGP model variance prediction in $\mathbb{R}$\\
  $w_\mathrm{n}$ & Weighting factor in the KDE estimation in $[0, 1]$ \\ 
  
  				 & \\
  
  $\alpha_{\boldsymbol{k}}$  & Coefficient of the PCE in $\mathbb{R}$ \\  
  $\alpha_i$ & Polynomial coefficient for the $i$-th polynomial in $\mathbb{R}$\\
  $\alpha_{l,\boldsymbol{k}}$ & Polynomial coefficient for the lengthscale parameter in $\mathbb{R}$\\
  $\alpha_{l,\boldsymbol{k},i}$ & Polynomial coefficient for the lengthscale parameter \\
                                & and $i$-th polynomial in $\mathbb{R}$\\
  $\boldsymbol{\alpha}_l$ & Polynomial coefficient vector for the \\
                          & lengthscale parameter in $\mathbb{R}^{N_{\alpha,l}}$\\
  $\alpha_{\sigma_\mathrm{n},\boldsymbol{k}}$ & Polynomial coefficient for the noise variance $\sigma_\mathrm{n}^2$ in $\mathbb{R}$\\
  $\alpha_{\sigma_\mathrm{n},\boldsymbol{k},i}$ & Polynomial coefficient for the noise variance $\sigma_\mathrm{n}^2$\\
                                                & and $i$-th polynomial in $\mathbb{R}$\\ 
  $\boldsymbol{\alpha}_{\sigma_\mathrm{n}}$ & Polynomial coefficient vector \\
                                            & for the noise variance $\sigma_\mathrm{n}^2$ in $\mathbb{R}^{N_{\alpha,\sigma_{\mathrm{n}}}}$\\
  $\beta$  & Parameter in Elastic Net that balances \\
           & the $l_1$ (Lasso) and $l_2$ (Ridge) penalties in $[0, 1]$ \\
  $\gamma_{\boldsymbol{k}}$  & Normalization constant associated with the multi-index $\boldsymbol{k}$ in $\mathbb{R}^{+}$ \\
  \vspace{-2mm}$\delta_{\boldsymbol{k}\boldsymbol{j}}$ & \vspace{-2mm}Kronecker delta \\
  $\boldsymbol{\epsilon}$ & Factors for data scaling in $\mathbb{R}^{2\times 1}$ \\
  $\zeta$ & Numerical stability term in $\mathbb{R}^+$ \\
  $\eta$ & Learning rate in $\mathbb{R}^+$ \\  
  $\vartheta$ & Quantile for separating the "better" and \\
              & "worse" groups in $(0, 1]$ \\ 
  $\boldsymbol{\theta}_i$ & Model parameters at iteration $i$ in $\mathbb{R}^{N_{\mathrm{\theta}}}$ \\  
  $\boldsymbol{\theta}$ & Hyperparameter vector of a GP in $\mathbb{R}^{n_x+2}$\\
  $\boldsymbol{\theta}_i$ & $i$-th Hyperparameter of a GP in $\mathbb{R}$\\
  $\boldsymbol{\theta}_\mathrm{PCEGP}$ &  Hyperparameter vector of the \\
                                       & PCEGP in $\mathbb{R}^{N_{\alpha,l}+ N_{\alpha,\sigma_{\mathrm{n}}} + n_k+1 \times 1}$\\  
  $\boldsymbol{\theta}_\mathrm{T}$ & Training hyperparameter vector of the in $\mathbb{R}^{6 \times 1}$\\  
  $\kappa_i$ & Smoothing coefficient for the first moment at iteration $i$ in $\mathbb{R}^+$ \\
  $\lambda$ & Regularization parameter in Elastic Net \\
            & that controls the overall strength \\
            & of the regularization in $\mathbb{R}^{+}$ \\
            
  $\mu$ & Mean prediction of a GP\\
  $\nu$ & Smoothing coefficient for the second moment in $\mathbb{R}^+$ \\
  $\xi_i$  & Single random input variable, where $i = 1, 2, \ldots, n_\xi$ in $\mathbb{R}$ \\
  $\boldsymbol{\xi}$ & Vector of random input variables, \\
                     & $\boldsymbol{\xi} = (\xi_1, \xi_2, \ldots, \xi_{n_\xi})$ in $\mathbb{R}^{n_\xi}$ \\  
  $\rho$ & Weight factor for fine-tuning between exploitation and exploration\\  
  $\sigma$ & Standard deviation in $\mathbb{R}^{+}$\\
  $\sigma^2$ & Variance prediction of a GP in $\mathbb{R}^{+}$\\
  $\sigma_\mathrm{f}^2$ & Output scale variance of a GP in $\mathbb{R}^+$\\
  $\boldsymbol{\sigma}_{\mathrm{f}}^2$ & Output scale variance vector of PCEGP in $\mathbb{R}^{n_k}$\\
  $\sigma_\mathrm{n}^2$ & Noise variance of a GP in $\mathbb{R}^+$\\ 
  $\hat{\sigma}^2_\mathrm{n}\left(x\right)$ & Prediction of the noise variance \\
                                            & of the scalar input $x$ by the PCE in $\mathbb{R}^{+}$\\
  $\hat{\sigma}_\mathrm{n}^2\left(\boldsymbol{x}\right)$ & Prediction of the noise variance \\
                                                         & of the input vector $\boldsymbol{x}$ by the PCE in $\mathbb{R}^{+}$\\
  $\hat{\boldsymbol{\sigma}}_\mathrm{n}^2\left(\boldsymbol{X}\right)$ & Prediction of the noise variance vector \\
                                                                      & by the PCE in $\mathbb{R}^{N \times N}$\\
  $\varphi$ & Cumulative distribution function\\
  $\phi_{k_i}(\xi_i)$ & Univariate orthogonal polynomials corresponding to the random variable $\xi_i$ with degree $k_i$ \\
  $\Phi_{\boldsymbol{k}}(\boldsymbol{\xi})$ & Multivariate polynomial basis functions in the PCE, \\
                                            & constructed as a product of univariate orthogonal polynomials \\
  $\psi_1$, $\psi_2$ & Parameters controlling the balance \\
                     & between exploration and exploitation \\
                     & with $\psi_1 \in (0, 1], \, \psi_2 \in (0, \sqrt{N}]$ \\ 
  \vspace{-2mm}$\Psi(N_{\mathrm{obs}})$ & \vspace{-2mm}Function for dynamically adjusting $\vartheta$ based on \\
                           & the number of observations in $\mathbb{N} \to (0, 1]$ \\
                           &\\
  
  $\mathcal{D}=\{\boldsymbol{x}_i, y_i\}_{i=1}^{N}$ & Training dataset with $N$ data points \\ 
  $\mathcal{D}_{\mathrm{HP}}$ & Set of previous observations of hyperparameters \\
                              & and their objective function values $\{(\boldsymbol{h}_i, o_{\mathrm{B},i})\}_{i=1}^N$ \\
  $\mathcal{D}_{\mathrm{HP}}(b)$ & "Better group": hyperparameters with better objective function values \\
  $\mathcal{D}_{\mathrm{HP}}(w)$ & "Worse group": hyperparameters with inferior objective function values \\
  $\mathcal{H}$ & Space of all possible hyperparameter configurations \\
  $\mathcal{K}$ & Set of all possible multi-indices $\boldsymbol{k}$ \\
  $\mathcal{K}_q$ & Set of multi-indices whose total degree \\
                  & does not exceed the maximum polynomial degree $q$ \\  
  $\mathcal{L}(\alpha)$ & Loss function in the Elastic Net optimization problem \\
  $\mathcal{N}$ & Normal distribution \\
  $\mathcal{T}$ & Transformation for PCE generated values \\
\end{longtable}

\section{Polynomial Chaos Expansion Bases}
\label{appendix_pce_bases}
\begin{figure}[t!]
    \centering
    \includegraphics[width=0.95\textwidth]{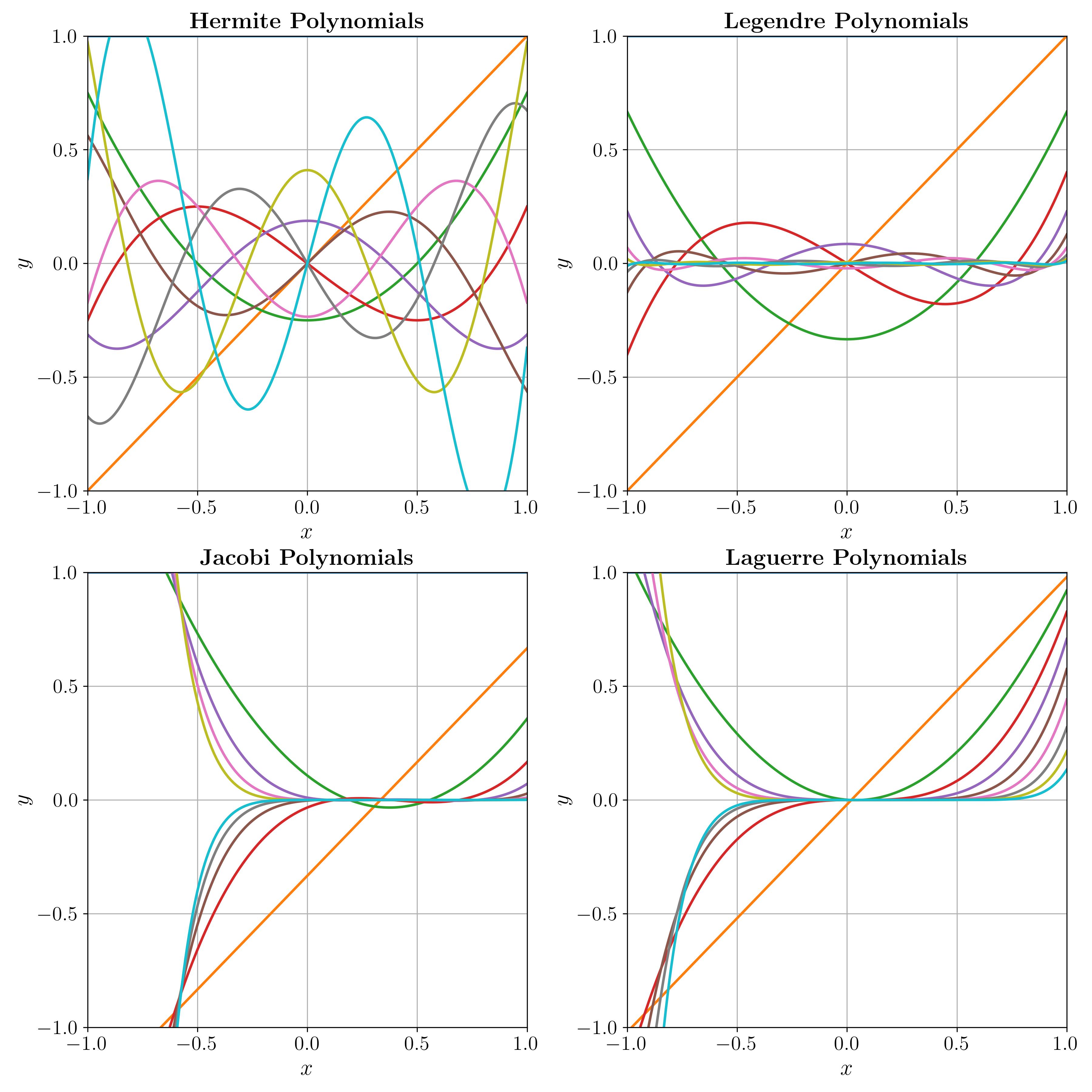}
    \caption{Polynomial bases for polynomial chaos expansion}
    \label{fig_pce_bases}
\end{figure}

In Table~\ref{tab_polynomial_bases}, the formulas of four different types of commonly used PCE bases are shown. In addition, the distribution and probability density function are shown for each PCE basis. In Figure~\ref{fig_pce_bases}, the function curves of ten polynomials for each polynomial basis are shown. In uncertainty quantification, these polynomial bases are commonly used to model uncertain parameters, depending on the distribution of data. All polynomial bases of the PCE are orthogonal to their probability density function and build computational efficient surrogates~\cite{Shen.2020}. In this work, the Legendre polynomials are used to calculate the hyperparameters of the GP. The concept of this work allows the combination of different polynomial bases to adapt the polynomials to better fit the model hyperparameters to the data. With each additional polynomial bases the number of adjustable polynomial coefficients increases, which makes it more difficult for the optimizer to find the optimum. The number of polynomial coefficients for the lengthscale parameter can be calculated using the formula $b \binom{n_x + q}{q}$, 
where $b$ denotes the number of different polynomial bases, $n_x$ represents the number of input variables, and $q$ is the polynomial degree.

\begin{table}[ht]
  \renewcommand{\arraystretch}{2}
  \centering
  \caption{Polynomial bases for polynomial chaos expansion}
  \begin{tabular}{llccc}
    \toprule
    Name & Distribution & Density function & Formula \\
    \midrule
    Hermite & Normal  & $\frac{1}{\sqrt{2\pi}}e^{-\frac{x^2}{2}}$ & $\mathrm{H}(x) = (-1)^ne^{\frac{x^2}{2}}\frac{\mathrm{d}^n}{\mathrm{d}x^n}(e^{-\frac{x^2}{2}})$ \\
    
    Legendre & Uniform & $\frac{1}{2}$ & $\mathrm{Le}_n(x) = \frac{1}{2^n}\binom{2n}{n}(1-x^2)^{n/2}$ \\
    
    Jacobi & Beta &  $\frac{(1-x)^\alpha(1+x)^\beta}{2^{\alpha+\beta+1}B(\alpha+1,\beta+1)}$ & $\mathrm{J}_n^{(\alpha, \beta)}(x) = \frac{1}{2^n n!} \sum_{k=0}^{n} (-1)^k \binom{n+\alpha}{k} \binom{n+\beta}{n-k}$ \\ 
    &&& $\hspace{4mm}\cdot (1-x)^{n-k} (1+x)^k$ \\
    
    Laguerre & Exponential &  $e^{-x}$ & $\mathrm{La}_n(x) = \frac{e^x}{n!}\frac{\mathrm{d}^n}{\mathrm{d}x^n}(x^ne^{-x})$ \\
    \bottomrule
  \end{tabular}
   \label{tab_polynomial_bases}
\end{table}
\newpage

\section{Transformations for Polynomial Chaos Expansion} \label{appendix_pce_transformations}
This section presents various transformations applied to a PCE to map the GP hyperparameters, specifically the lengthscale \(\hat{l}(x)\) and the noise variance \(\hat{\sigma}_\mathrm{n}^{2}(x)\), to the positive real domain \(\mathbb{R}^{+}\). The PCE is constructed using Legendre polynomials based on a uniformly distributed input variable within the interval \([-1, 1]\) and is generated with a polynomial degree of 10. Each polynomial term is multiplied by randomly generated coefficients within the range \([-1, 1]\).

\begin{figure}[b!]
    \centering
    \includegraphics[width=0.95\textwidth]{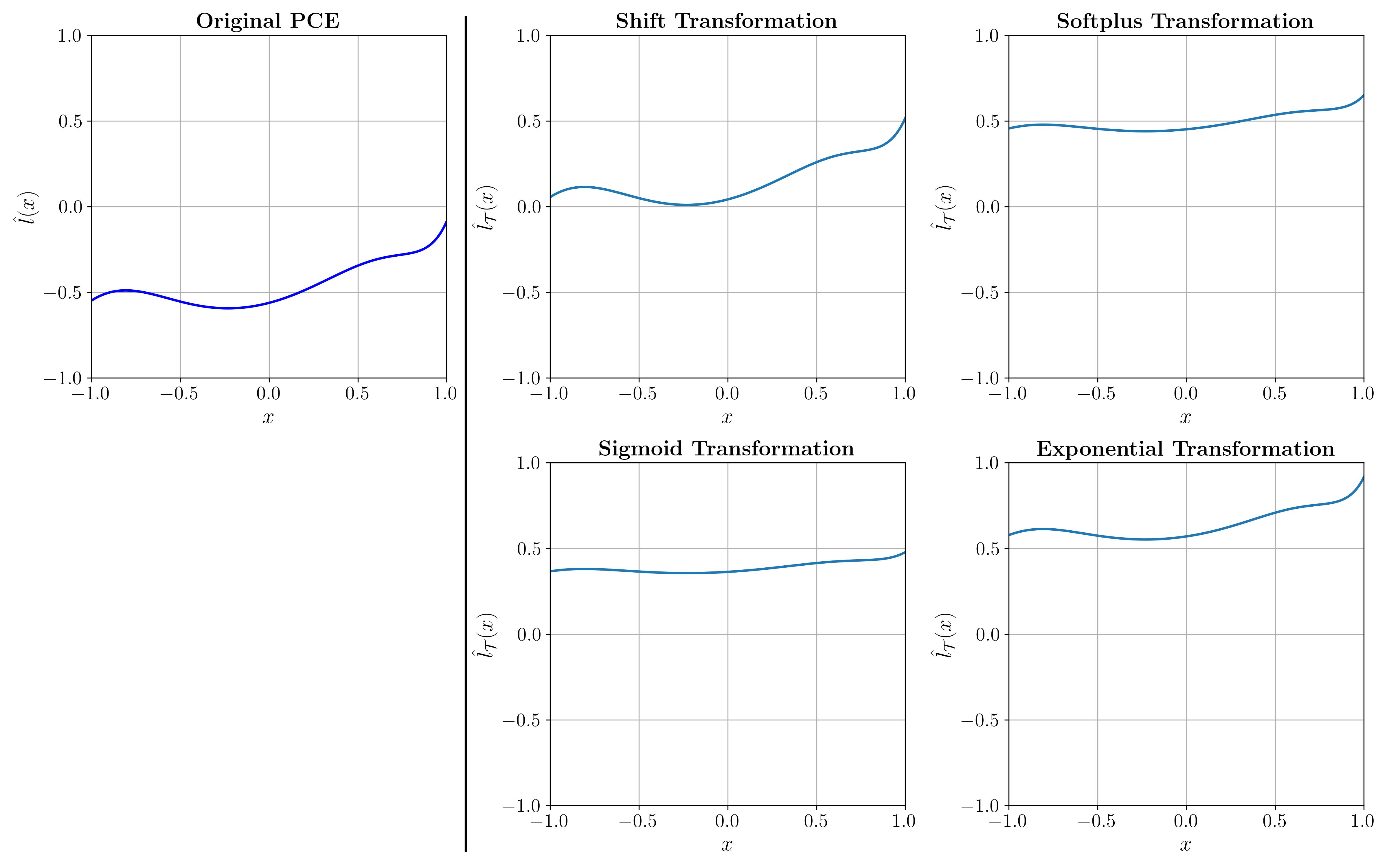}
    \caption{Transformations for polynomial chaos expansion generated values}
    \label{fig_pce_transformations}
\end{figure}

The PCE transformation is demonstrated for the lengthscale parameter $\hat{l}\left(x\right)$, while the same approach can be applied to the noise parameter $\hat{\sigma}_{\mathrm{n}}(x)$. The PCE output, denoted by $\hat{l}(x)$, and the transformed PCE outputs, denoted by $\hat{l}_{\mathcal{T}}(x)$, are shown in Figure~\ref{fig_pce_transformations}. The original PCE output $\hat{l}(x)$ is shown in the top-left plot, while the remaining plots show the transformed PCE values $\hat{l}_{\mathcal{T}}(x)$. These transformations are chosen to map the output to $\mathbb{R}^{+}$, with each transformation having different characteristics.

The shift transformation adjusts $\hat{l}(x)$ by adding a constant offset to ensure that the minimum value is set to $0.01$ whenever any negative values are generated by the PCE, thereby guaranteeing positive outputs. However, this approach is computationally challenging as it requires continuously computing $\min(\hat{l}(x))$ to determine the necessary shift.
 The softplus transformation, given by $\log\left(1 + \mathrm{e}^{\hat{l}(x)}\right)$, provides a smooth approximation of the ReLU function, making all values of $\hat{l}_{\mathcal{T}}(x)$ positive while introducing a smoothing effect. The sigmoid transformation fits $\hat{l}(x)$ into a bounded range of $\left[0, 1\right]$, using the function $\frac{1}{1 + e^{-\hat{l}(x)}}$. Finally, the exponential transformation, represented by $\mathrm{e}^{\hat{l}(x)}$, increases the growth rate of larger $\hat{l}(x)$ values, generating exponentially scaled outputs for $\hat{l}_{\mathcal{T}}(x)$.

The applied transformations and their formulas are listed in Table \ref{tab_pce_transformations}, which highlights the nonlinear mappings applied to $\hat{l}(x)$. These transformations ensure that the GP hyperparameters $\hat{l}(x)$ and $\hat{\sigma}_\mathrm{n}^{2}(x)$ are appropriately scaled within the positive real domain, which is crucial for stable model training and inference.
\begin{table}[h!]
	\renewcommand{\arraystretch}{2}
	\caption{Transformations applied to the PCE for GP hyperparameters}
	\label{tab_pce_transformations}
    \centering
    \begin{tabular}{lc}
        \toprule
        Transformation & Formula \\
        \midrule
        Shift & $\hat{l}_{\mathcal{T}}(x) = \hat{l}(x) + (0.01 - \min(\hat{l}(x)))$ if $\min(\hat{l}(x)) < 0$  \\
      
        Softplus & $\hat{l}_{\mathcal{T}}(x) = \log(1 + \mathrm{e}^{\hat{l}(x)})$ \\
        
        Sigmoid & $\hat{l}_{\mathcal{T}}(x) = \frac{1}{1 + e^{-\hat{l}(x)}}$ \\
        
        Exponential & $\hat{l}_{\mathcal{T}}(x) = e^{\hat{l}(x)}$ \\
        \bottomrule
    \end{tabular}
\end{table}

\newpage
\section{Stationary and Non-stationary Covariance Functions}
\label{appendix_covariance_functions}
\begin{figure}[h!]
    \centering
    \includegraphics[width=\textwidth]{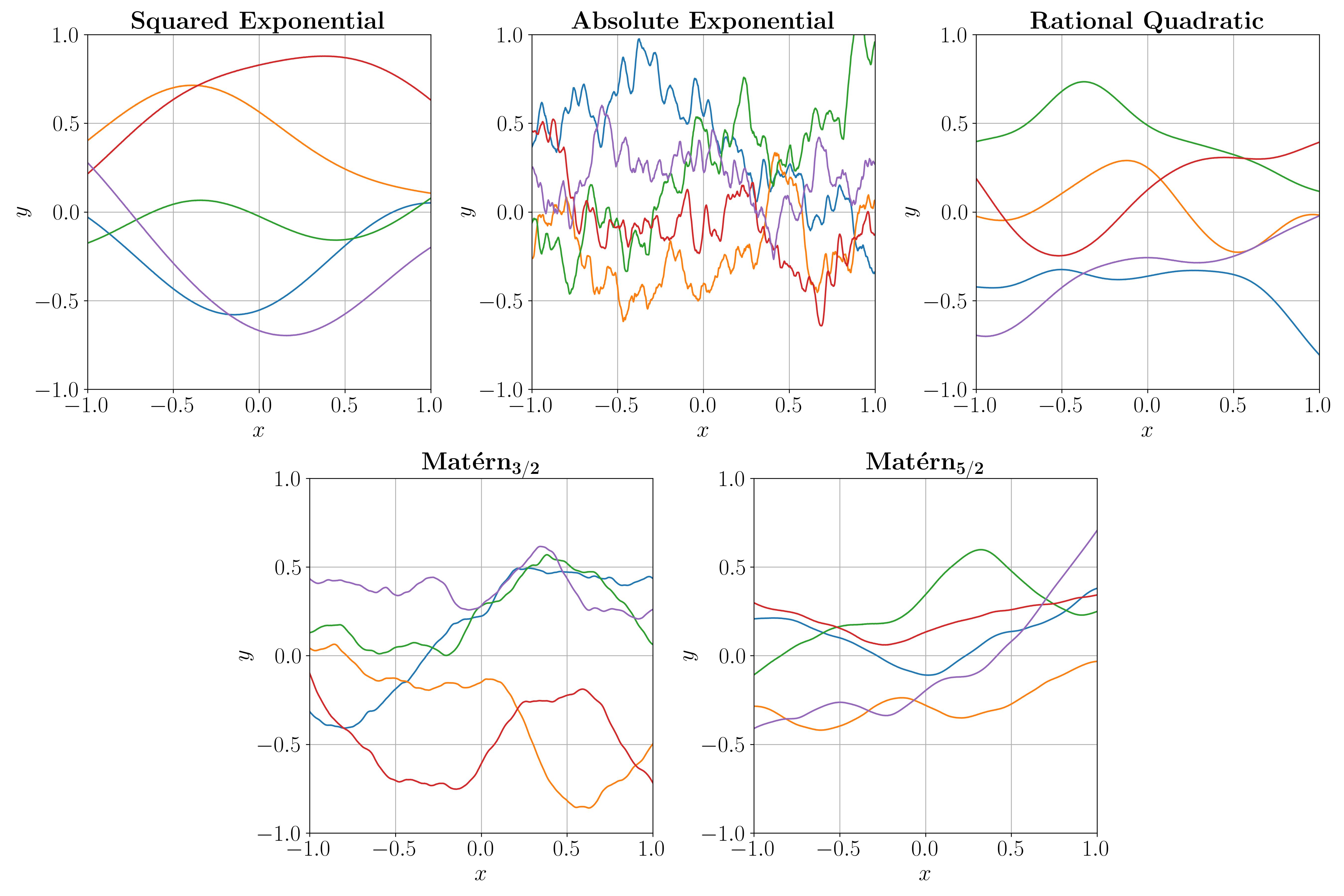}
    \caption{Covariance functions for Gaussian processes}
    \label{fig_covars}
\end{figure}

The following Table~\ref{tab_stationary_covariance_functions} shows five commonly used stationary covariance functions~\cite{Rasmussen.2006}. These five covariance functions are used in this work, which can represent a wide range of functional curves and thus ensure the variability of the model. These chosen covariance functions are the squared exponential, absolute exponential, Matérn$_{3/2}$, Matérn$_{5/2}$, and rational quadratic. This covers function curves from very smooth to very wiggly. In Table~\ref{tab_nonstationary_covariance_functions}, the non-stationary covariance versions of Table~\ref{tab_stationary_covariance_functions} are shown, which are also used in~\cite{Cremanns.2017}. Here, the lengthscale parameter $l\left(\boldsymbol{x}\right)$ depends directly on the input $\boldsymbol{x}$. In Figure~\ref{fig_covars}, the function curve of the covariance functions are shown. These function curves are random drawn from the prior distribution of the GP with a zero mean. 

\begin{table}[h!]
	\renewcommand{\arraystretch}{2}
	\caption{Stationary covariance functions}
	\label{tab_stationary_covariance_functions}
    \centering
    \begin{tabular}{lc}
        \toprule
        Name & Formula \\
        \midrule
        Squared exponential & $k(\boldsymbol{x}, \boldsymbol{x}') = \sigma_\mathrm{f}^2 \exp\left(-\frac{\|\boldsymbol{x} - \boldsymbol{x}'\|^2}{2\,l^2}\right)$ \\
      
        Absolute exponential & $k(\boldsymbol{x}, \boldsymbol{x}') = \sigma_\mathrm{f}^2 \exp\left(-\frac{\|\boldsymbol{x} - \boldsymbol{x}'\|}{l}\right)$ \\
        
        Matérn$_{3/2}$ & $k(\boldsymbol{x}, \boldsymbol{x}') = \sigma_\mathrm{f}^2 \left(1 + \frac{\sqrt{3}\|\boldsymbol{x} - \boldsymbol{x}'\|}{l}\right)\exp\left(-\frac{\sqrt{3}\|\boldsymbol{x} - \boldsymbol{x}'\|}{l}\right)$ \\
        
        Matérn$_{5/2}$ & $k(\boldsymbol{x}, \boldsymbol{x}') = \sigma_\mathrm{f}^2 \left(1 + \frac{\sqrt{5}\|\boldsymbol{x} - \boldsymbol{x}'\|}{l} + \frac{5\|\boldsymbol{x} - \boldsymbol{x}'\|^2}{3l^2}\right)\exp\left(-\frac{\sqrt{5}\|\boldsymbol{x} - \boldsymbol{x}'\|}{l}\right)$ \\
        
        Rational quadratic & $k(\boldsymbol{x}, \boldsymbol{x}') = \sigma_\mathrm{f}^2 \left(1 + \frac{\|\boldsymbol{x} - \boldsymbol{x}'\|^2}{2\,\alpha\, l^2}\right)^{-\alpha}$ \\
        \bottomrule
    \end{tabular}
\end{table}

\begin{table}[h!]
	\renewcommand{\arraystretch}{2}
	\caption{Non-stationary covariance functions}
	\label{tab_nonstationary_covariance_functions}
    \centering
    \begin{tabular}{lc}
        \toprule
        Name & Formula \\
        \midrule
        Squared exponential & $k(\boldsymbol{x}, \boldsymbol{x}') = \sigma_\mathrm{f}^2 \exp\left(-\frac{\|l\left(\boldsymbol{x}\right)\,\boldsymbol{x} - l\left(\boldsymbol{x}'\right)\,\boldsymbol{x}'\|^2}{2}\right)$ \\
        
        Absolute exponential & $k(\boldsymbol{x}, \boldsymbol{x}') = \sigma_\mathrm{f}^2 \exp\left(-\|l\left(\boldsymbol{x}\right)\,\boldsymbol{x} - l\left(\boldsymbol{x}'\right)\, \boldsymbol{x}'\|\right)$ \\
        
        Matérn$_{3/2}$ & $k(\boldsymbol{x}, \boldsymbol{x}') = \sigma_\mathrm{f}^2 \left(1 + \sqrt{3}\,\|l\left(\boldsymbol{x}\right)\,\boldsymbol{x} - l\left(\boldsymbol{x}'\right)\,\boldsymbol{x}'\|\right)$ \\
        &$\hspace{14mm}\cdot\exp\left(-\sqrt{3}\,\|l\left(\boldsymbol{x}\right)\,\boldsymbol{x} - \l\left(\boldsymbol{x}'\right)\,\boldsymbol{x}'\|\right)$ \\
        
        Matérn$_{5/2}$ & $k(\boldsymbol{x}, \boldsymbol{x}') = \sigma_\mathrm{f}^2 \left(1 + \sqrt{5}\,\|l\left(\boldsymbol{x}\right)\,\boldsymbol{x} - l\left(\boldsymbol{x}'\right)\,\boldsymbol{x}'\| + \frac{5\|l\left(\boldsymbol{x}\right)\,\boldsymbol{x} - l\left(\boldsymbol{x}'\right)\,\boldsymbol{x}'\|^2}{3}\right)$\\
        &  $\hspace{-14mm}\cdot\exp\left(-\sqrt{5}\,\|l\left(\boldsymbol{x}\right)\,\boldsymbol{x} - l\left(\boldsymbol{x}'\right)\,\boldsymbol{x}'\|\right)$ \\

        Rational quadratic & $k(\boldsymbol{x}, \boldsymbol{x}') = \sigma_\mathrm{f}^2 \left(1 + \frac{\|l\left(\boldsymbol{x}\right)\,\boldsymbol{x} -l\left(\boldsymbol{x}'\right)\, \boldsymbol{x}'\|^2}{2\alpha}\right)^{-\alpha}$ \\
        \bottomrule
    \end{tabular}
\end{table}

\newpage

\newpage

\section{Algorithm for Model Prediction}\label{appendix_model_prediction}
The following Algorithm~\ref{pcegp_prediction} describes how the prediction of the PCEGP model works. The input is the test input data $\boldsymbol{x}^*$, which represents the point at which the prediction is to be made. In the model, the covariance function $k_\Sigma \left(\boldsymbol{x},\boldsymbol{x}'\right)$, the polynomials for lengthscale $\hat{l}_i\left(\boldsymbol{x}\right)$, and the polynomials for noise variance $\hat{\sigma}_\mathrm{n}^2\left(\boldsymbol{x}\right)$ are specified. Additionally, the covariance matrix depending on the training data $\boldsymbol{K}\left( \boldsymbol{X}_\mathrm{s}, \hat{\boldsymbol{l}}\left(\boldsymbol{X}_\mathrm{s}\right)\right) + \hat{\boldsymbol{\sigma}}_\mathrm{n}^2\left(\boldsymbol{X}_\mathrm{s}\right)\,\boldsymbol{I}$ and the model hyperparameters $\boldsymbol{\theta}_\mathrm{PCEGP} = \left[\begin{array}{c c c c c} q_l & q_{\sigma_\mathrm{n}} & \boldsymbol{\alpha}_l^{\intercal} & \boldsymbol{\alpha}_{\sigma_\mathrm{n}}^{\intercal} & \boldsymbol{\sigma}_{\mathrm{f}}^2 \end{array} \right]^{\intercal}$ are part of the trained model and are required to predict the output $\hat{\boldsymbol{y}}_\mathrm{PCEGP}$, consisting of the predicted mean and uncertainty. The polynomial degree for the PCE is given by $q_l$ for the lengthscale and $q_{\sigma_\mathrm{n}}$ for the noise variance. The corresponding polynomial coefficients are represented by the vectors $\boldsymbol{\alpha}_l$ and $\boldsymbol{\alpha}_{\sigma_{\mathrm{n}}}$, respectively. The output scale, denoted by $\sigma_{\mathrm{f}}^2$, defines the overall variance of the model prediction.

First, the input data point $\boldsymbol{x}^*$ is scaled to the corresponding scaled input $\boldsymbol{x}_\mathrm{s}^*$. For each of the covariance functions, the lengthscale parameters $\hat{l}(\boldsymbol{x})$ are calculated as a function of the input data using the polynomial chaos expansion (PCE). Next, the covariance $k_\Sigma(\boldsymbol{x}_\mathrm{s},\boldsymbol{x}_\mathrm{s}^*)$ between the previously observed training data points $\mbox{$\mathcal{D}=\{\boldsymbol{x}_{\mathrm{s},i}\}_{i=1}^{N}$}$ and the scaled input point $\boldsymbol{x}_\mathrm{s}^*$ is computed. This results in the covariance vector $\boldsymbol{k}_\Sigma\left(\boldsymbol{X}_\mathrm{s}, \boldsymbol{x}_\mathrm{s}^*\right)$.

Subsequently, the point-dependent noise variance $\hat{\sigma}_\mathrm{n}^2\left(\boldsymbol{x}_\mathrm{s}^*\right)$ is computed using the PCE coefficients. The scaled output mean $\hat{y}_\mathrm{PCEGP,s,m}$ is then computed by multiplying the covariance vector with the inverse of the covariance matrix and scaled training outputs. The scaled output variance $\hat{y}_\mathrm{PCEGP,s,v}$ is also calculated, incorporating the noise variance.

Finally, the scaled predictions are rescaled back to the original value range, resulting in the final predicted mean $\hat{y}_\mathrm{PCEGP,m}$ and variance $\hat{y}_\mathrm{PCEGP,v}$ for the PCEGP model, which are summarized in $\hat{\boldsymbol{y}}_\mathrm{PCEGP}$.

\begin{algorithm}[htb!]
   \caption{PCEGP model prediction}
   \label{pcegp_prediction}
\begin{algorithmic}
   \STATE {\bfseries Input:} Input data $\boldsymbol{x}^*$
   \STATE {\bfseries Output:} Model prediction $\hat{\boldsymbol{y}}_\mathrm{PCEGP}$
   \STATE {\bfseries Specified in the model:} Covariance function $k_\Sigma \left(\boldsymbol{x},\boldsymbol{x}'\right)$, polynomials for lengthscale $\hat{l}_i\left(\boldsymbol{x}\right)$, polynomials for noise variance $\hat{\sigma}_\mathrm{n}^2\left(\boldsymbol{x}\right)$, covariance matrix $\boldsymbol{K}\left( \boldsymbol{X}_\mathrm{s}, \hat{\boldsymbol{l}}\left(\boldsymbol{X}_\mathrm{s}\right)\right) + \hat{\boldsymbol{\sigma}}_\mathrm{n}^2\left(\boldsymbol{X}_\mathrm{s}\right)\,\boldsymbol{I}$, and model hyperparameters $\boldsymbol{\theta}_\mathrm{PCEGP} = \left[\begin{array}{c c c c c} q_l & q_{\sigma_{\mathrm{n}}} & \boldsymbol{\alpha}_l^{\intercal} & \boldsymbol{\alpha}_{\sigma_{\mathrm{n}}}^{\intercal} & \boldsymbol{\sigma}_{\mathrm{f}}^2\end{array} \right]^{\intercal}$
\end{algorithmic}
\begin{algorithmic}[1]
   \STATE $\boldsymbol{x}_\mathrm{s}^* \leftarrow$ Scale input data $\boldsymbol{x}^*$
   \FOR {$i = 1 \dots n_k$}
      \STATE $\hat{l}_i\left(\boldsymbol{x}_\mathrm{s}^*\right) \leftarrow \sum_{\boldsymbol{k} \in \mathcal{K}_l} \alpha_{l,i,\boldsymbol{k}}\,\Phi_{\boldsymbol{k}}\left(\boldsymbol{x}_\mathrm{s}^*\right)$
   \ENDFOR
   \FOR {$i = 1 \dots N$}
      \STATE $k_\Sigma(\boldsymbol{x}_{\mathrm{s},i},\boldsymbol{x}_\mathrm{s}^*) \leftarrow \left\|\hat{l}\left(\boldsymbol{x}_{\mathrm{s},i}\right)\, \boldsymbol{x}_{\mathrm{s},i} - \hat{l}\left(\boldsymbol{x}_\mathrm{s}^*\right)\, \boldsymbol{x}_\mathrm{s}^*\right\|$
   \ENDFOR
   \STATE $\boldsymbol{k}_\Sigma\left(\boldsymbol{X}_\mathrm{s}, \boldsymbol{x}_\mathrm{s}^*\right) \leftarrow \left[ k_\Sigma(\boldsymbol{x}_{\mathrm{s},1},\boldsymbol{x}_\mathrm{s}^*), \dots, k_\Sigma(\boldsymbol{x}_{\mathrm{s},N},\boldsymbol{x}_\mathrm{s}^*) \right]^\intercal$
   \STATE $\hat{\sigma}_\mathrm{n}^2\left(\boldsymbol{x}_\mathrm{s}^*\right) \leftarrow \sum_{\boldsymbol{k} \in \mathcal{K}_{\sigma_\mathrm{n}}} \alpha_{\sigma_\mathrm{n},\boldsymbol{k}}\,\Phi_{\boldsymbol{k}}\left(\boldsymbol{x}_\mathrm{s}^*\right)$
   \STATE $\hat{y}_\mathrm{PCEGP,s,m} \leftarrow \boldsymbol{k}_\Sigma\left(\boldsymbol{X}_\mathrm{s}, \boldsymbol{x}_\mathrm{s}^* \right)^\intercal \left( \boldsymbol{K}\left( \boldsymbol{X}_\mathrm{s}, \hat{\boldsymbol{l}}\left(\boldsymbol{X}_\mathrm{s}\right)\right) + \hat{\boldsymbol{\sigma}}_\mathrm{n}^2\left(\boldsymbol{X}_\mathrm{s}\right) \boldsymbol{I} \right)^{-1} \boldsymbol{y}$
   \STATE $\hat{y}_\mathrm{PCEGP,s,v} \leftarrow k_\Sigma\left(\boldsymbol{x}_\mathrm{s}^*, \boldsymbol{x}_\mathrm{s}^* \right) + \hat{\sigma}_\mathrm{n}^2\left(\boldsymbol{x}_\mathrm{s}^*\right) - \boldsymbol{k}_\Sigma\left(\boldsymbol{X}_\mathrm{s}, \boldsymbol{x}_\mathrm{s}^* \right)^\intercal \left( \boldsymbol{K}\left( \boldsymbol{X}_\mathrm{s}, \hat{\boldsymbol{l}}\left(\boldsymbol{X}_\mathrm{s}\right)\right) + \hat{\boldsymbol{\sigma}}_\mathrm{n}^2\left(\boldsymbol{X}_\mathrm{s}\right) \boldsymbol{I} \right)^{-1} \boldsymbol{k}_\Sigma\left(\boldsymbol{X}_\mathrm{s}, \boldsymbol{x}_\mathrm{s}^*\right)$
   \STATE $\hat{\boldsymbol{y}}_\mathrm{PCEGP} \leftarrow$ Rescale the output $\boldsymbol{y}_\mathrm{PCEGP,s}$ of the PCEGP model
\end{algorithmic}
\end{algorithm}

\clearpage

\section{Algorithm for Hyperparameter Optimization}
\label{appendix_hpo}
The following Algorithm~\ref{pcegp_algorithm_nadam} outlines the hyperparameter optimization process of the PCEGP model. The input for the hyperparameter optimization includes the data $\mathcal{D}=\{\boldsymbol{X}, \boldsymbol{y}\}$, where $\boldsymbol{X}$ represents the input data, and $\boldsymbol{y}$ represents the corresponding output values of $N$ data points. Additionally, the input parameters are the number of trials $n_\mathrm{T}$, the number of initial trials $n_\mathrm{IT}$, and the number of folds $n_\mathrm{F}$ for cross-validation.

The number of initial trials $n_\mathrm{IT}$ defines the number of hyperparameter optimization runs performed using random search. For the remaining $n_\mathrm{T} - n_\mathrm{IT}$ trials, the Tree-structured Parzen estimator (TPE) algorithm is used.  The $n_\mathrm{F}$ parameter specifies the number of folds used in $k$-fold cross-validation, where the dataset is split into $n_\mathrm{F}$ subsets, with each subset used once as the validation set, while the remaining data are used for training.

The output of the hyperparameter optimization is the best set of model hyperparameters $\boldsymbol{\theta}_\mathrm{PCEGP} = \left[\begin{array}{c c c c c} q_l & q_{\sigma_\mathrm{n}} & \boldsymbol{\alpha}_l^{\intercal} & \boldsymbol{\alpha}_{\sigma_{\mathrm{n}}}^{\intercal} & \boldsymbol{\sigma}_{\mathrm{f}}^2 \end{array} \right]^{\intercal}$. Here, $q_l$ and $q_{\sigma_\mathrm{n}}$ represent the polynomial degree for the PCE for the lengthscale and noise variance. The vectors $\boldsymbol{\alpha}_l$ and $\boldsymbol{\alpha}_{\sigma_{\mathrm{n}}}$ denote the polynomial coefficients for the lengthscale and noise variance, respectively. The output scale is denoted by $\sigma_{\mathrm{f}}^2$ and determines the overall variance of the model prediction. The training hyperparameters learning rate $\eta$, number of gradient descent iterations $n_\mathrm{I}$, regularization balancing term $\beta$, regularization term $\lambda$, and data scaling parameters $\boldsymbol{\epsilon}$ are also optimized in this process and summarized in $\boldsymbol{\theta}_\mathrm{T} \in \mathbb{R}^{6\times 1}$.

The algorithm proceeds as follows. First, the training hyperparameters are suggested. If the condition $trial \leq n_\mathrm{IT}$ holds, the hyperparameters are selected through random search. Otherwise, the TPE algorithm is applied to suggest the hyperparameters $q_l$, $q_{\sigma_\mathrm{n}}$, $\eta$, $N_\mathrm{I}$, $\beta$, $\lambda$, and $\boldsymbol{\epsilon}$. 

Next, the input and output data are scaled. For each of the $n_\mathrm{F}$ folds in the $k$-fold cross-validation, the training and validation datasets are split. The PCEGP model is then set up, with the covariance function $k_\Sigma \left(\boldsymbol{x},\boldsymbol{x}'\right)$, the polynomial for the lengthscale $\hat{l}\left(\boldsymbol{x}\right)$, and the polynomial for the noise variance $\hat{\sigma}_\mathrm{n}^2\left(\boldsymbol{x}\right)$ specified. 

For each iteration $i = 1 \dots n_\mathrm{I}$, the model parameters $\boldsymbol{\theta}_{\mathrm{PCEGP}}$ are optimized by minimizing the negative log-likelihood (NLL) loss $\mathcal{L}_{\mathrm{NLL}}\left(\boldsymbol{\theta}_{\mathrm{PCEGP},i}, \mathcal{D}_{\mathrm{train}}\right)$ plus an elastic net regularization term $\mathcal{L}_{\mathrm{EN}}\left(\boldsymbol{\theta}_{\mathrm{PCEGP},i}\right)$ using the Nadam optimization method. The gradient of the loss function with respect to the model parameters, $\nabla_{\boldsymbol{\theta}} \left(\mathcal{L}_{\mathrm{NLL}}(\boldsymbol{\theta}_{\mathrm{PCEGP},i}, \mathcal{D}_{\mathrm{train}}) + \mathcal{L}_{\mathrm{EN}}(\boldsymbol{\theta}_{\mathrm{PCEGP},i})\right)$, is calculated for each step. The first and second moment estimates, $m_i$ and $v_i$, are updated based on the smoothing factor $\kappa_i$ and the gradient $\boldsymbol{g}_i$, and subsequently bias-corrected to obtain $\hat{m}_i$ and $\hat{v}_i$. The model parameters $\boldsymbol{\theta}_{\mathrm{PCEGP},i}$ are then updated using $\hat{m}_i$, $\hat{v}_i$, the learning rate $\eta$, and the numerical stability term $\zeta$, ensuring robust and efficient optimization.

After each fold, the validation loss $\mathcal{L}_{\mathrm{val}}\left(\boldsymbol{\theta}_{\mathrm{PCEGP},i}, \mathcal{D}_{\mathrm{val}}\right)$ is computed, and the average validation loss $\overline{\mathcal{L}}_{\mathrm{val}}$ is determined by averaging across all folds. If the average validation loss $\overline{\mathcal{L}}_{\mathrm{val}}$ improves, the best model hyperparameters $\boldsymbol{\theta}_{\mathrm{PCEGP},\mathrm{best}}$ and training hyperparameters $\eta_{\mathrm{best}}$, $N_\mathrm{I, best}$, $\beta_{\mathrm{best}}$, $\lambda_{\mathrm{best}}$, and $\boldsymbol{\epsilon}_{\mathrm{best}}$ are updated. This process repeats until the number of trials $n_\mathrm{T}$ is reached.
\newpage
\begin{algorithm}[H]
   \caption{Hyperparameter Optimization of PCEGP Model using Nadam}
   \label{pcegp_algorithm_nadam}
\begin{algorithmic}
   \STATE {\bfseries Input:} Data $\mathcal{D}=\{\boldsymbol{X}, \boldsymbol{y}\}$, number of trials $n_\mathrm{T}$, number of initial trials $n_\mathrm{IT}$, number of folds $n_\mathrm{F}$, number of gradient descent iterations $n_\mathrm{I}$
   \STATE {\bfseries Output:} Optimized model hyperparameters $\boldsymbol{\theta}_\mathrm{PCEGP} = \left[\begin{array}{c c c c c} q_l & q_{\sigma_\mathrm{n}} & \boldsymbol{\alpha}_l^{\intercal} & \boldsymbol{\alpha}_{\sigma_{\mathrm{n}}}^{\intercal} & \boldsymbol{\sigma}_{\mathrm{f}}^2 \end{array} \right]^{\intercal}$ and training hyperparameters $\eta$, $n_\mathrm{I}$, $\beta$, $\lambda$, $\boldsymbol{\epsilon}$
   \STATE {\bfseries Specified:} Covariance function $k_\Sigma \left(\boldsymbol{x},\boldsymbol{x}'\right)$, polynomials for lengthscale $\hat{l}\left(\boldsymbol{x}\right)$ and noise variance $\hat{\sigma}_\mathrm{n}^2\left(\boldsymbol{x}\right)$
\end{algorithmic}
\begin{algorithmic}[1]
   \REPEAT
   \STATE Suggest hyperparameters:
   \IF{$trial \leq n_\mathrm{IT}$}
      \STATE Random Search for $q_l$, $q_{\sigma_\mathrm{n}}$, $\eta$, $n_\mathrm{I}$, $\lambda$, $\boldsymbol{\epsilon}$
   \ELSE
      \STATE TPE for $q_l$, $q_{\sigma_\mathrm{n}}$, $\eta$, $n_\mathrm{I}$, $\lambda$, $\boldsymbol{\epsilon}$
   \ENDIF
   \STATE $\boldsymbol{X}_\mathrm{s}, \boldsymbol{y}_\mathrm{s} \leftarrow$ Scale input and output data $\boldsymbol{X}, \boldsymbol{y}$ with scaling parameters $\boldsymbol{\epsilon}$ on $\mathcal{D}_{\mathrm{train}}$
   \FOR{$fold = 1 \dots n_\mathrm{F}$}
      \STATE $\mathcal{D}_{\mathrm{train}}, \mathcal{D}_{\mathrm{val}} \leftarrow$ k-fold split of data $\mathcal{D}$
      \STATE Define covariance function $k_\Sigma \left(\boldsymbol{x},\boldsymbol{x}'\right)$, polynomials $\hat{l}\left(\boldsymbol{x}\right)$ and $\hat{\sigma}_\mathrm{n}^2\left(\boldsymbol{x}\right)$
      \FOR{$i = 1 \dots N_\mathrm{I}$}
         \STATE $\boldsymbol{g}_i \leftarrow \nabla_{\boldsymbol{\theta}_{\mathrm{PCEGP}}} \mathcal{L}(\boldsymbol{\theta}_{\mathrm{PCEGP},i-1})$
         \STATE $m_i \leftarrow \kappa_i \, m_{i-1} + (1 - \kappa_i) \, g_i$
         \STATE $v_i \leftarrow \nu \, v_{i-1} + (1 - \nu) \, g_i^2$
         \STATE $\hat{m}_i \leftarrow \frac{\kappa_{i+1}\, m_i}{1 - \prod_{j=1}^{i+1} \kappa_j} + \frac{(1 - \kappa_i)\, g_i}{1 - \prod_{j=1}^i \kappa_j}$ 
         \STATE $\hat{v}_i \leftarrow \frac{v_i}{1 - \nu^i}$
         \STATE $\boldsymbol{\theta}_{\mathrm{PCEGP},i} \leftarrow \boldsymbol{\theta}_{\mathrm{PCEGP},i-1} - \frac{\eta}{\sqrt{\hat{v}_i} + \zeta} \,\hat{m}_i$
      \ENDFOR
      \STATE $\mathcal{L}_{\mathrm{val}} \leftarrow \mathcal{L}_{\mathrm{val}}\left(\boldsymbol{\theta}_{\mathrm{PCEGP}}, \mathcal{D}_{\mathrm{val}}\right)$
   \ENDFOR
   \STATE $\overline{\mathcal{L}}_{\mathrm{val}} \leftarrow \frac{1}{n_\mathrm{F}} \sum_{fold=1}^{n_\mathrm{F}} \mathcal{L}_{\mathrm{val}}\left(\boldsymbol{\theta}_{\mathrm{PCEGP}}, \mathcal{D}_{\mathrm{val}}\right)$
   \IF{$\overline{\mathcal{L}}_{\mathrm{val}} < \overline{\mathcal{L}}_{\mathrm{val,best}}$}
      \STATE $\boldsymbol{\theta}_{\mathrm{PCEGP},\mathrm{best}} \leftarrow \boldsymbol{\theta}_{\mathrm{PCEGP}}$
      \STATE $\eta_{\mathrm{best}}, n_\mathrm{I, best}, \lambda_{\mathrm{best}}, \boldsymbol{\epsilon}_{\mathrm{best}} \leftarrow \eta, n_\mathrm{I}, \lambda, \boldsymbol{\epsilon}$
   \ENDIF
   \STATE $trial \leftarrow trial + 1$
   \UNTIL{$trial = n_\mathrm{T}$}
\end{algorithmic}
\end{algorithm}

\newpage

\section{PCEGP Model Setup for Toy Examples}\label{pcegp_structure_toy_example}
In this section, the PCEGP model setup for toy examples is described, which is shown in Figure~\ref{fig_pcegp_toy_examples}. The inputs are scaled with the min-max-scaler in the range of $\left[-0.5, 0.5\right]$. The output is normalized. To compute the input-dependent lengthscale parameters $\hat{l}\left(\boldsymbol{x}^*\right)$, the Legendre PCE basis is used, with a fixed polynomial degree $q_l=10$ and a uniform distribution between the range of $\left[-1, 1\right]$. To ensure that lengthscale the parameters $\hat{l}\left(\boldsymbol{x}^*\right)$ remain positive, the softplus transformation is applied. The squared exponential covariance function is used to calculate the elements of the covariance matrix. The noise variance is calculated homoscedastic.
\begin{figure*}[h!]
  \centering
  \includegraphics[width=\textwidth]{./Figure11.eps} 
  \caption{Polynomial Chaos Expanded Gaussian Process setup for toy examples}
  \label{fig_pcegp_toy_examples}
\end{figure*}

\section{Error Metrics for Model Evaluation}\label{error_metrics}
Several common error metrics are used to evaluate the effectiveness of different ML models. These metrics used in this work provide a quantitative assessment of the prediction accuracy and reliability of each model, allowing for an objective comparison. Below are descriptions of six key metrics: Mean Absolute Error (MAE), Median Absolute Error (MedAE), Mean Squared Error (MSE), Root Mean Squared Error (RMSE), R\(^2\) score, and Test Negative Log-Likelihood (NLL).

The \textbf{Mean Absolute Error (MAE)} is defined as
\[
\text{MAE} = \frac{1}{M} \sum_{i=1}^{M} |y_i - \hat{y}_i|.
\]
The MAE represents the average absolute difference between the predicted values \(\hat{y}_i\) and the true values \(y_i\). Since it does not square the errors, MAE is less sensitive to large deviations compared to other metrics like MSE. The resulting error is expressed in the same units as the target variable \(y_i\), making it intuitive to interpret.

The \textbf{Median Absolute Error (MedAE)} is given by
\[
\text{MedAE} = \text{median}(|y_i - \hat{y}_i|).
\]
This metric calculates the median of the absolute errors, providing a robust alternative to MAE by reducing the influence of extreme outliers. MedAE is useful when error distributions are non-symmetric or contain large values, and, like MAE, is expressed in the same units as \(y_i\).

The \textbf{Mean Squared Error (MSE)} is calculated as
\[
\text{MSE} = \frac{1}{M} \sum_{i=1}^{M} (y_i - \hat{y}_i)^2.
\]
The MSE measures the average squared deviation between predicted and true values, heavily penalizing larger errors due to the squaring operation. This makes it sensitive to outliers, as these contribute disproportionately to the metric. Unlike the MAE, the MSE has squared units of the target variable, which can make direct interpretation more challenging.

The \textbf{Root Mean Squared Error (RMSE)} is derived from MSE and is given by
\[
\text{RMSE} = \sqrt{\frac{1}{M} \sum_{i=1}^{M} (y_i - \hat{y}_i)^2}.
\]
The RMSE provides an error metric that is in the same units as the target variable $y$, which makes it more interpretable than MSE. Since it is the square root of the MSE, the RMSE is sensitive to larger errors but does so in a way that remains interpretable in the original units.

The \textbf{R\(^2\) score} or \textbf{coefficient of determination} is calculated as
\[
R^2 = 1 - \frac{\sum_{i=1}^{M} (y_i - \hat{y}_i)^2}{\sum_{i=1}^{M} (y_i - \bar{y})^2}.
\]
This metric measures the proportion of variance in the true values \(y_i\) explained by the model. An R\(^2\) score of 1 indicates a perfect fit, while scores close to zero imply that the model provides little predictive power beyond a simple mean-based approach. Negative values indicate that the model performs worse than a model that predicts the mean value for all instances.

The \textbf{Test Negative Log-Likelihood (NLL)}, particularly useful for probabilistic models, is given by
\[
\text{NLL} = \frac{1}{M} \sum_{i=1}^{M} \left( 0.5 \frac{(y_i - \mu_i)^2}{\sigma_i^2} + 0.5 \log(2 \pi \sigma_i^2) \right).
\]
Here, \(\mu_{i}\) and \(\sigma_{i}^2\) are the predicted mean and variance, respectively, for each instance. The first term, \(0.5 \frac{(y_i - \mu_{i})^2}{\sigma_{i}^2}\), penalizes the model for inaccurate predictions, with larger penalties for underconfident or overconfident predictions. The second term, \(0.5 \log(2 \pi \sigma_{i}^2)\), represents the uncertainty associated with the predictions, rewarding models that have accurate estimates of their uncertainty. Lower NLL values indicate better model performance and improved alignment between the model's confidence and actual prediction errors.

These metrics collectively allow for a thorough evaluation of model performance, considering both prediction accuracy and uncertainty.

\newpage

\section{PCEGP Model Setup for Benchmark Experiments}\label{pcegp_structure_experiments}
In this section, the PCEGP model setup for experiments is described. For all experiments, the same model setup is used, which is shown in Figure~\ref{fig_pcegp_experiments}. Here, the input scaling is part of HPO. The output is normalized, which proved to be the best configuration. To compute the input-dependent lengthscale parameters $\hat{l}\left(\boldsymbol{x}^*\right)$, the Legendre PCE basis is used for each covariance function, with a fixed polynomial degree $q_l=5$ and a uniform distribution between the range of $\left[-1, 1\right]$. 

For the UCI datasets, preliminary tests have shown that no HP transformation achieves the best results. Therefore, the hyperparameter transformation is omitted in the final model setup for experiments. With the squared exponential, absolute exponential, Matérn$_{3/2}$, Matérn$_{5/2}$, and rational quadratic, five different covariance functions are used. These covariance functions vary from very sharp variations to smooth variations and are be combined to map a wide range of function curves. For the modeling tasks in this work, the noise variance is homoscedastic.

\begin{figure*}[h!]
  \centering
  \includegraphics[width=\textwidth]{./Figure12.eps} 
  \caption{Polynomial Chaos Expanded Gaussian Process setup for experiments}
  \label{fig_pcegp_experiments}
\end{figure*}

\newpage

\section{UCI Datasets}\label{uci_datasets}

The datasets used in this work are obtained from the UCI machine learning repository\footnote{\url{https://archive.ics.uci.edu/ml/index.php}}. They include input variables ranging between five and 13 and data points varying between 103 and $6,497$. This diversity enables a comprehensive evaluation of the models across different dataset structures. A brief description of these datasets is provided below.

\textbf{Boston Housing:} This dataset contains 506 data points, with 13 input variables and one output variable. The output represents the median value of owner-occupied homes in Boston, aiming to predict housing prices based on various factors like crime rate, average number of rooms, and proximity to employment centers.  
\textbf{Inputs:} Crime rate, proportion of residential land zoned for large lots, proportion of non-retail business acres, Charles River dummy variable, nitric oxides concentration, average number of rooms, age, distance to employment centers, access to highways, property-tax rate, pupil-teacher ratio, black population proportion, lower status of the population  
\textbf{Output:} Median value of owner-occupied homes  

\textbf{Concrete Compressive Strength:} Comprising $1,030$ instances, this dataset includes eight input variables and one output variable. It is used to model the compressive strength of concrete as a function of its material composition, such as cement content, water, and age of the concrete.  
\textbf{Inputs:} Cement, slag, fly ash, water, superplasticizer, coarse aggregate, fine aggregate, age  
\textbf{Output:} Compressive strength of concrete  

\textbf{Energy Efficiency:} This dataset consists of 768 data points with eight input features and two output variables. It models the energy efficiency of building heating and cooling loads, based on factors like wall area, roof area, and glazing area.  
\textbf{Inputs:} Relative compactness, surface area, wall area, roof area, overall height, orientation, glazing area, glazing area distribution  
\textbf{Outputs:} Heating load, cooling load  

\textbf{Airfoil Self-Noise:} This dataset has $1,503$ observations, with five input variables and one output variable. It seeks to model the noise level generated by an airfoil in a wind tunnel, based on parameters such as frequency, angle of attack, and airspeed.  
\textbf{Inputs:} Frequency, angle of attack, chord length, free-stream velocity, suction side displacement thickness  
\textbf{Output:} Sound pressure level

\textbf{Yacht Hydrodynamics:} Consisting of 308 data points, this dataset includes six input variables and one output variable. The model predicts the hydrodynamic resistance of a yacht based on hull dimensions and speed, making it useful in naval architecture.  
\textbf{Inputs:} Longitudinal position of the center of buoyancy, prismatic coefficient, length-displacement ratio, beam-draught ratio, length-beam ratio, Froude number  
\textbf{Output:} Residuary resistance per unit weight of displacement  

\textbf{Wine Quality:} This dataset contains $4,898$ data points for white wine and $1,599$ data points for red wine, totaling $6,497$ observations. Each sample has eleven input features and one output variable, which represents the quality score of the wine, evaluated on a scale from 0 to 10. Features include attributes like acidity, sugar content, and alcohol level, which influence the perceived quality.  
\textbf{Inputs:} Fixed acidity, volatile acidity, citric acid, residual sugar, chlorides, free sulfur dioxide, total sulfur dioxide, density, pH, sulphates, alcohol  
\textbf{Output:} Quality score

\textbf{Concrete Slump Test:} This dataset consists of 103 data points, each with seven input variables and three output variables. It is used to predict the behavior of concrete mixtures in terms of slump flow, strength, and flow spread, which are critical for construction and material science. The dataset helps optimize concrete mix designs to achieve desired properties.  
\textbf{Inputs:} Cement, slag, fly ash, water, superplasticizer, coarse aggregate, fine aggregate  
\textbf{Outputs:} Slump flow, strength, flow spread

\section{Results on UCI Datasets}\label{appendix_results}
In the following Table~\ref{tab:model_performance}, the results of the model performances are shown. For model evaluation, five different error metrics are used. The best error metrics are highlighted in dark green and the worst in green-yellow. The darker the shade of green, the better the error metric. If no color is highlighted, there is no convergence for the model on the corresponding dataset. In Table~\ref{tab:performance_legend}, the colors represent the relative performance of the models, ranging from \textit{Best} for the model with the highest performance (leftmost cell) to \textit{Worst} for the model with the lowest performance (rightmost cell). Intermediate colors represent performances ranked between these extremes.

\begin{table}[h!]
\centering
\caption{Performance Legend}
\begin{tabular}{p{1.7cm} p{1.7cm} p{1.7cm} p{1.7cm} p{1.7cm} p{1.7cm} }
\ApplyAbsoluteGradient{1}{Best} & 
\ApplyAbsoluteGradient{2}{} & 
\ApplyAbsoluteGradient{3}{} & 
\ApplyAbsoluteGradient{4}{} & 
\ApplyAbsoluteGradient{5}{} & 
\ApplyAbsoluteGradient{6}{Worst} \\
\end{tabular}
\label{tab:performance_legend}
\end{table}

\begin{landscape}
\begin{longtable}{l r r r r r r}
\caption{Results of the model performances on the UCI datasets}
    \label{tab:model_performance} \\
    \toprule
    \textbf{Dataset / Metric} & \textbf{GP} & \textbf{GP ARD} & \textbf{DGP} & \textbf{GPBoost} & \textbf{HHK GP} & \textbf{PCEGP} \\
    \midrule
    \endfirsthead

    \toprule
    \textbf{Dataset / Metric} & \textbf{GP} & \textbf{GP ARD} & \textbf{DGP} & \textbf{GPBoost} & \textbf{HHK GP} & \textbf{PCEGP} \\
    \midrule
    \endhead

    \midrule
    \multicolumn{7}{r}{\textit{Continued on next page}} \\
    \midrule
    \endfoot

    \bottomrule
    \endlastfoot

\multicolumn{7}{l}{\textbf{Boston Housing}} \\
MAE    & \ApplyAbsoluteGradient{6}{$9.231 \pm 0.043$} 
       & \ApplyAbsoluteGradient{3}{$2.108 \pm 0.033$} 
       & \ApplyAbsoluteGradient{4}{$2.143 \pm 0.070$} 
       & \ApplyAbsoluteGradient{5}{$2.407 \pm 0.038$} 
       & \ApplyAbsoluteGradient{1}{$1.997 \pm 0.081$}
       & \ApplyAbsoluteGradient{2}{$2.024 \pm 0.053$} \\
MedAE  & \ApplyAbsoluteGradient{4}{$1.556 \pm 0.049$} 
       & \ApplyAbsoluteGradient{3}{$1.492 \pm 0.044$} 
       & \ApplyAbsoluteGradient{5}{$1.560 \pm 0.040$} 
       & \ApplyAbsoluteGradient{6}{$1.726 \pm 0.041$} 
       & \ApplyAbsoluteGradient{1}{$1.372 \pm 0.059$}
       & \ApplyAbsoluteGradient{2}{$1.427 \pm 0.076$} \\
MSE    & \ApplyAbsoluteGradient{6}{$10.381 \pm 0.271$} 
       & \ApplyAbsoluteGradient{5}{$9.698 \pm 0.346$} 
       & \ApplyAbsoluteGradient{4}{$10.108 \pm 0.737$} 
       & \ApplyAbsoluteGradient{3}{$12.883 \pm 0.643$} 
       & \ApplyAbsoluteGradient{2}{$9.3606 \pm 1.150$} 
       & \ApplyAbsoluteGradient{1}{$8.936 \pm 0.579$} \\
RMSE   & \ApplyAbsoluteGradient{6}{$3.158 \pm 0.046$} 
       & \ApplyAbsoluteGradient{5}{$3.047 \pm 0.053$} 
       & \ApplyAbsoluteGradient{4}{$3.081 \pm 0.087$} 
       & \ApplyAbsoluteGradient{3}{$3.492 \pm 0.099$} 
       & \ApplyAbsoluteGradient{2}{$2.979 \pm 0.173$} 
       & \ApplyAbsoluteGradient{1}{$2.930 \pm 0.051$} \\
R2     & \ApplyAbsoluteGradient{4}{$0.871 \pm 0.008$} 
       & \ApplyAbsoluteGradient{3}{$0.879 \pm 0.008$} 
       & \ApplyAbsoluteGradient{5}{$0.869 \pm 0.024$} 
       & \ApplyAbsoluteGradient{6}{$0.838 \pm 0.013$} 
       & \ApplyAbsoluteGradient{2}{$0.882 \pm 0.017$} 
       & \ApplyAbsoluteGradient{1}{$0.887 \pm 0.010$} \\
NLL    & \ApplyAbsoluteGradient{5}{$5.724 \pm 0.036$}
	   & \ApplyAbsoluteGradient{2}{$2.684 \pm 0.006$}
	   & \ApplyAbsoluteGradient{6}{$7.346 \pm 0.104$}
	   & \ApplyAbsoluteGradient{3}{$2.780 \pm 0.045$}
	   & \ApplyAbsoluteGradient{1}{$2.508 \pm 0.065$}
	   & \ApplyAbsoluteGradient{4}{$3.767 \pm 0.007$}\\
\midrule

\multicolumn{7}{l}{\textbf{Concrete Compressive}} \\
MAE    & \ApplyAbsoluteGradient{6}{$17.771 \pm 0.032$} 
       & \ApplyAbsoluteGradient{5}{$18.051 \pm 0.035$} 
       & \ApplyAbsoluteGradient{4}{$4.112 \pm 0.149$} 
       & \ApplyAbsoluteGradient{3}{$2.838 \pm 0.047$} 
       & \ApplyAbsoluteGradient{1}{$2.317 \pm 0.064$} 
       & \ApplyAbsoluteGradient{2}{$2.401 \pm 0.065$} \\
MedAE  & \ApplyAbsoluteGradient{6}{$3.796 \pm 0.033$} 
       & \ApplyAbsoluteGradient{5}{$3.158 \pm 0.057$} 
       & \ApplyAbsoluteGradient{4}{$3.107 \pm 0.172$} 
       & \ApplyAbsoluteGradient{3}{$1.956 \pm 0.068$} 
       & \ApplyAbsoluteGradient{2}{$1.409 \pm 0.039$} 
       & \ApplyAbsoluteGradient{1}{$1.340 \pm 0.038$} \\
MSE    & \ApplyAbsoluteGradient{6}{$41.975 \pm 0.554$} 
       & \ApplyAbsoluteGradient{5}{$31.110 \pm 0.515$} 
       & \ApplyAbsoluteGradient{4}{$30.769 \pm 1.793$} 
       & \ApplyAbsoluteGradient{3}{$17.583 \pm 0.869$} 
       & \ApplyAbsoluteGradient{1}{$14.311 \pm 0.802$} 
       & \ApplyAbsoluteGradient{2}{$17.034 \pm 1.185$} \\
RMSE   & \ApplyAbsoluteGradient{6}{$6.456 \pm 0.045$} 
       & \ApplyAbsoluteGradient{5}{$5.556 \pm 0.045$} 
       & \ApplyAbsoluteGradient{4}{$5.509 \pm 0.169$} 
       & \ApplyAbsoluteGradient{3}{$4.149 \pm 0.104$} 
       & \ApplyAbsoluteGradient{1}{$3.723 \pm 0.105$} 
       & \ApplyAbsoluteGradient{2}{$4.054 \pm 0.129$} \\
R2     & \ApplyAbsoluteGradient{6}{$0.847 \pm 0.003$} 
       & \ApplyAbsoluteGradient{5}{$0.886 \pm 0.003$} 
       & \ApplyAbsoluteGradient{4}{$0.888 \pm 0.007$} 
       & \ApplyAbsoluteGradient{3}{$0.936 \pm 0.003$} 
       & \ApplyAbsoluteGradient{1}{$0.948 \pm 0.003$} 
       & \ApplyAbsoluteGradient{2}{$0.938 \pm 0.005$} \\
NLL    & \ApplyAbsoluteGradient{4}{$6.453 \pm 0.014$}
	   & \ApplyAbsoluteGradient{5}{$6.616 \pm 0.015$}
	   & \ApplyAbsoluteGradient{6}{$7.606 \pm 0.295$}
	   & \ApplyAbsoluteGradient{1}{$2.881 \pm 0.020$}
	   & \ApplyAbsoluteGradient{2}{$2.672 \pm 0.030$}
	   & \ApplyAbsoluteGradient{3}{$4.467 \pm 0.002$}\\

\midrule

\multicolumn{7}{l}{\textbf{Yacht}} \\
MAE    & \ApplyAbsoluteGradient{6}{$13.817 \pm 0.095$} 
       & \ApplyAbsoluteGradient{5}{$13.594 \pm 0.087$} 
       & \ApplyAbsoluteGradient{4}{$0.445 \pm 0.012$} 
       & \ApplyAbsoluteGradient{3}{$0.324 \pm 0.028$} 
       & \ApplyAbsoluteGradient{1}{$0.196 \pm 0.025$} 
       & \ApplyAbsoluteGradient{2}{$0.219 \pm 0.011$} \\
MedAE  & \ApplyAbsoluteGradient{6}{$2.173 \pm 0.048$} 
       & \ApplyAbsoluteGradient{5}{$0.735 \pm 0.033$} 
       & \ApplyAbsoluteGradient{4}{$0.291 \pm 0.011$} 
       & \ApplyAbsoluteGradient{3}{$0.134 \pm 0.010$} 
       & \ApplyAbsoluteGradient{1}{$0.044 \pm 0.006$} 
       & \ApplyAbsoluteGradient{2}{$0.105 \pm 0.008$} \\
MSE    & \ApplyAbsoluteGradient{6}{$22.765 \pm 0.628$} 
       & \ApplyAbsoluteGradient{5}{$6.814 \pm 0.308$} 
       & \ApplyAbsoluteGradient{4}{$0.582 \pm 0.037$} 
       & \ApplyAbsoluteGradient{3}{$0.466 \pm 0.132$} 
       & \ApplyAbsoluteGradient{2}{$0.278 \pm 0.083$} 
       & \ApplyAbsoluteGradient{1}{$0.175 \pm 0.002$} \\
RMSE   & \ApplyAbsoluteGradient{6}{$4.632 \pm 0.045$} 
       & \ApplyAbsoluteGradient{5}{$2.474 \pm 0.040$} 
       & \ApplyAbsoluteGradient{4}{$0.709 \pm 0.022$} 
       & \ApplyAbsoluteGradient{3}{$0.609 \pm 0.059$} 
       & \ApplyAbsoluteGradient{2}{$0.455 \pm 0.069$} 
       & \ApplyAbsoluteGradient{1}{$0.392 \pm 0.025$} \\
R2     & \ApplyAbsoluteGradient{6}{$0.894 \pm 0.008$} 
       & \ApplyAbsoluteGradient{5}{$0.970 \pm 0.002$} 
       & \ApplyAbsoluteGradient{4}{$0.997 \pm 0.000$} 
       & \ApplyAbsoluteGradient{3}{$0.998 \pm 0.001$} 
       & \ApplyAbsoluteGradient{2}{$0.999 \pm 0.001$} 
       & \ApplyAbsoluteGradient{1}{$0.999 \pm 0.000$} \\
NLL    & \ApplyAbsoluteGradient{3}{$6.350 \pm 0.043$}
	   & \ApplyAbsoluteGradient{4}{$6.422 \pm 0.039$}
	   & \ApplyAbsoluteGradient{6}{$46.554 \pm 0.410$}
	   & \ApplyAbsoluteGradient{2}{$0.865 \pm 0.079$}
	   & \ApplyAbsoluteGradient{1}{$-0.643 \pm 0.215$}
	   & \ApplyAbsoluteGradient{5}{$34.981 \pm 21.954$}\\
\midrule

\multicolumn{7}{l}{\textbf{Energy Heating Load}} \\
MAE    & \ApplyAbsoluteGradient{6}{$11.083 \pm 0.020$} 
       & \ApplyAbsoluteGradient{5}{$11.099 \pm 0.017$} 
       & \ApplyAbsoluteGradient{3}{$0.522 \pm 0.040$} 
       & \ApplyAbsoluteGradient{4}{$0.561 \pm 0.039$} 
       & \ApplyAbsoluteGradient{1}{$0.200 \pm 0.007$} 
       & \ApplyAbsoluteGradient{2}{$0.289 \pm 0.008$} \\
MedAE  & \ApplyAbsoluteGradient{6}{$1.056 \pm 0.030$} 
       & \ApplyAbsoluteGradient{5}{$0.928 \pm 0.021$} 
       & \ApplyAbsoluteGradient{3}{$0.382 \pm 0.034$} 
       & \ApplyAbsoluteGradient{4}{$0.425 \pm 0.044$} 
       & \ApplyAbsoluteGradient{1}{$0.126 \pm 0.005$} 
       & \ApplyAbsoluteGradient{2}{$0.193 \pm 0.006$} \\
MSE    & \ApplyAbsoluteGradient{6}{$4.761 \pm 0.105$} 
       & \ApplyAbsoluteGradient{5}{$2.471 \pm 0.084$} 
       & \ApplyAbsoluteGradient{3}{$0.521 \pm 0.069$} 
       & \ApplyAbsoluteGradient{4}{$0.587 \pm 0.070$} 
       & \ApplyAbsoluteGradient{1}{$0.107 \pm 0.011$} 
       & \ApplyAbsoluteGradient{2}{$0.190 \pm 0.012$} \\
RMSE   & \ApplyAbsoluteGradient{6}{$2.170 \pm 0.022$} 
       & \ApplyAbsoluteGradient{5}{$1.555 \pm 0.023$} 
       & \ApplyAbsoluteGradient{3}{$0.700 \pm 0.050$} 
       & \ApplyAbsoluteGradient{4}{$0.751 \pm 0.044$} 
       & \ApplyAbsoluteGradient{1}{$0.316 \pm 0.015$} 
       & \ApplyAbsoluteGradient{2}{$0.429 \pm 0.013$} \\
R2     & \ApplyAbsoluteGradient{6}{$0.952 \pm 0.001$} 
       & \ApplyAbsoluteGradient{5}{$0.975 \pm 0.001$} 
       & \ApplyAbsoluteGradient{3}{$0.995 \pm 0.001$} 
       & \ApplyAbsoluteGradient{4}{$0.994 \pm 0.001$} 
       & \ApplyAbsoluteGradient{1}{$0.999 \pm 0.000$} 
       & \ApplyAbsoluteGradient{2}{$0.998 \pm 0.000$} \\
NLL    & \ApplyAbsoluteGradient{4}{$6.435 \pm 0.013$}
	   & \ApplyAbsoluteGradient{3}{$6.416 \pm 0.012$}
	   & \ApplyAbsoluteGradient{6}{$40.127 \pm 1.332$}
	   & \ApplyAbsoluteGradient{2}{$1.143 \pm 0.059$}
	   & \ApplyAbsoluteGradient{1}{$0.621 \pm 0.249$}
	   & \ApplyAbsoluteGradient{5}{$19.976 \pm 38.420$}\\
\midrule
\multicolumn{7}{l}{\textbf{Energy Cooling Load}} \\
MAE    & \ApplyAbsoluteGradient{6}{$10.506 \pm 0.020$} 
       & \ApplyAbsoluteGradient{5}{$10.581 \pm 0.020$} 
       & \ApplyAbsoluteGradient{3}{$1.168 \pm 0.011$} 
       & \ApplyAbsoluteGradient{2}{$0.799 \pm 0.064$} 
       & \ApplyAbsoluteGradient{1}{$0.313 \pm 0.018$} 
       & \ApplyAbsoluteGradient{2}{$0.445 \pm 0.019$} \\
MedAE  & \ApplyAbsoluteGradient{5}{$0.697 \pm 0.026$} 
       & \ApplyAbsoluteGradient{2}{$0.284 \pm 0.012$} 
       & \ApplyAbsoluteGradient{4}{$0.620 \pm 0.014$} 
       & \ApplyAbsoluteGradient{6}{$0.689 \pm 0.079$} 
       & \ApplyAbsoluteGradient{1}{$0.196 \pm 0.015$} 
       & \ApplyAbsoluteGradient{3}{$0.308 \pm 0.013$} \\
MSE    & \ApplyAbsoluteGradient{5}{$2.101 \pm 0.061$} 
       & \ApplyAbsoluteGradient{3}{$0.445 \pm 0.048$} 
       & \ApplyAbsoluteGradient{6}{$2.997 \pm 0.049$} 
       & \ApplyAbsoluteGradient{4}{$1.036 \pm 0.121$} 
       & \ApplyAbsoluteGradient{1}{$0.272 \pm 0.045$} 
       & \ApplyAbsoluteGradient{2}{$0.426 \pm 0.060$} \\
RMSE   & \ApplyAbsoluteGradient{6}{$1.439 \pm 0.020$} 
       & \ApplyAbsoluteGradient{3}{$0.656 \pm 0.032$} 
       & \ApplyAbsoluteGradient{5}{$1.721 \pm 0.015$} 
       & \ApplyAbsoluteGradient{4}{$0.999 \pm 0.065$} 
       & \ApplyAbsoluteGradient{1}{$0.502 \pm 0.037$} 
       & \ApplyAbsoluteGradient{2}{$0.637 \pm 0.040$} \\
R2     & \ApplyAbsoluteGradient{5}{$0.977 \pm 0.001$} 
       & \ApplyAbsoluteGradient{3}{$0.995 \pm 0.001$} 
       & \ApplyAbsoluteGradient{6}{$0.967 \pm 0.001$} 
       & \ApplyAbsoluteGradient{4}{$0.988 \pm 0.001$} 
       & \ApplyAbsoluteGradient{1}{$0.997 \pm 0.001$} 
       & \ApplyAbsoluteGradient{2}{$0.995 \pm 0.001$} \\
NLL    & \ApplyAbsoluteGradient{5}{$50.292 \pm 0.154$}
	   & \ApplyAbsoluteGradient{6}{$243.651 \pm 1.954$}
	   & \ApplyAbsoluteGradient{3}{$16.024 \pm 0.074$}
	   & \ApplyAbsoluteGradient{2}{$1.435 \pm 0.085$}
	   & \ApplyAbsoluteGradient{4}{$31.004 \pm 2.048$}
	   & \ApplyAbsoluteGradient{1}{$0.458 \pm 0.063$}\\
\midrule
\multicolumn{7}{l}{\textbf{Airfoil Self Noise}} \\
MAE    & \ApplyAbsoluteGradient{6}{$7.252 \pm 0.008$} 
       & \ApplyAbsoluteGradient{5}{$7.381 \pm 0.005$} 
       & \ApplyAbsoluteGradient{4}{$1.252 \pm 0.019$} 
       & \ApplyAbsoluteGradient{3}{$1.247 \pm 0.033$} 
       & \ApplyAbsoluteGradient{2}{$0.757 \pm 0.082$} 
       & \ApplyAbsoluteGradient{1}{$0.681 \pm 0.061$} \\
MedAE  & \ApplyAbsoluteGradient{6}{$1.646 \pm 0.021$} 
       & \ApplyAbsoluteGradient{5}{$1.336 \pm 0.012$} 
       & \ApplyAbsoluteGradient{4}{$0.922 \pm 0.021$} 
       & \ApplyAbsoluteGradient{3}{$0.897 \pm 0.027$} 
       & \ApplyAbsoluteGradient{2}{$0.491 \pm 0.072$} 
       & \ApplyAbsoluteGradient{1}{$0.437 \pm 0.013$} \\
MSE    & \ApplyAbsoluteGradient{6}{$8.473 \pm 0.113$} 
       & \ApplyAbsoluteGradient{5}{$6.182 \pm 0.100$} 
       & \ApplyAbsoluteGradient{3}{$2.976 \pm 0.088$} 
       & \ApplyAbsoluteGradient{4}{$3.173 \pm 0.127$} 
       & \ApplyAbsoluteGradient{2}{$1.479 \pm 0.419$} 
       & \ApplyAbsoluteGradient{1}{$1.133 \pm 0.139$} \\
RMSE   & \ApplyAbsoluteGradient{6}{$2.902 \pm 0.018$} 
       & \ApplyAbsoluteGradient{5}{$2.476 \pm 0.017$} 
       & \ApplyAbsoluteGradient{3}{$1.717 \pm 0.025$} 
       & \ApplyAbsoluteGradient{4}{$1.769 \pm 0.034$} 
       & \ApplyAbsoluteGradient{2}{$1.1411 \pm 0.108$} 
       & \ApplyAbsoluteGradient{1}{$1.046 \pm 0.047$} \\
R2     & \ApplyAbsoluteGradient{6}{$0.819 \pm 0.003$} 
       & \ApplyAbsoluteGradient{5}{$0.868 \pm 0.002$} 
       & \ApplyAbsoluteGradient{3}{$0.937 \pm 0.002$} 
       & \ApplyAbsoluteGradient{4}{$0.932 \pm 0.003$} 
       & \ApplyAbsoluteGradient{2}{$0.968 \pm 0.009$} 
       & \ApplyAbsoluteGradient{1}{$0.976 \pm 0.003$} \\
NLL    & \ApplyAbsoluteGradient{4}{$4.868 \pm 0.007$}
	   & \ApplyAbsoluteGradient{5}{$5.436 \pm 0.006$}
	   & \ApplyAbsoluteGradient{6}{$9.640 \pm 0.191$}
	   & \ApplyAbsoluteGradient{2}{$2.107 \pm 0.026$}
	   & \ApplyAbsoluteGradient{1}{$1.442 \pm 0.060$}
	   & \ApplyAbsoluteGradient{3}{$3.740 \pm 0.057$}\\
\midrule

\multicolumn{7}{l}{\textbf{Wine Quality Red}} \\
MAE    & \ApplyAbsoluteGradient{2}{$0.473 \pm 0.001$} 
       & \ApplyAbsoluteGradient{3}{$0.476 \pm 0.002$} 
       & \ApplyAbsoluteGradient{4}{$0.486 \pm 0.001$} 
       & \multirow{6}{*}{No convergence} 
       & \ApplyAbsoluteGradient{5}{$0.418 \pm 0.078$} 
       & \ApplyAbsoluteGradient{1}{$0.373 \pm 0.006$} \\
MedAE  & \ApplyAbsoluteGradient{2}{$0.366 \pm 0.004$} 
       & \ApplyAbsoluteGradient{3}{$0.374 \pm 0.004$} 
       & \ApplyAbsoluteGradient{4}{$0.388 \pm 0.003$} 
       &  
       & \ApplyAbsoluteGradient{5}{$0.258 \pm 0.030$} 
       & \ApplyAbsoluteGradient{1}{$0.216 \pm 0.011$} \\
MSE    & \ApplyAbsoluteGradient{2}{$0.385 \pm 0.002$} 
       & \ApplyAbsoluteGradient{3}{$0.389 \pm 0.002$} 
       & \ApplyAbsoluteGradient{4}{$0.395 \pm 0.002$} 
       &  
       & \ApplyAbsoluteGradient{5}{$0.595 \pm 0.655$} 
       & \ApplyAbsoluteGradient{1}{$0.333 \pm 0.007$} \\
RMSE   & \ApplyAbsoluteGradient{2}{$0.619 \pm 0.002$} 
       & \ApplyAbsoluteGradient{3}{$0.623 \pm 0.002$} 
       & \ApplyAbsoluteGradient{4}{$0.628 \pm 0.002$} 
       &  
       & \ApplyAbsoluteGradient{5}{$0.646 \pm 0.160$} 
       & \ApplyAbsoluteGradient{1}{$0.578 \pm 0.006$} \\
R2     & \ApplyAbsoluteGradient{2}{$0.406 \pm 0.004$} 
       & \ApplyAbsoluteGradient{3}{$0.399 \pm 0.004$} 
       & \ApplyAbsoluteGradient{4}{$0.389 \pm 0.004$} 
       &  
       & \ApplyAbsoluteGradient{5}{$0.016 \pm 0.184$} 
       & \ApplyAbsoluteGradient{1}{$0.484 \pm 0.010$} \\
NLL    & \ApplyAbsoluteGradient{2}{$1.443 \pm 0.003$}
       & \ApplyAbsoluteGradient{3}{$1.451 \pm 0.002$}
       & \ApplyAbsoluteGradient{1}{$1.377 \pm 0.002$}
       &  
       & \ApplyAbsoluteGradient{5}{$8,792 \pm 18,881$}
       & \ApplyAbsoluteGradient{4}{$1.522 \pm 0.002$}\\

\multicolumn{7}{l}{\textbf{Wine Quality White}} \\
MAE    & \ApplyAbsoluteGradient{2}{$0.501 \pm 0.001$} 
       & \ApplyAbsoluteGradient{3}{$0.520 \pm 0.001$} 
       & \ApplyAbsoluteGradient{4}{$0.544 \pm 0.001$} 
       & \multirow{6}{*}{No convergence} 
       & \multirow{6}{*}{No convergence} 
       & \ApplyAbsoluteGradient{1}{$0.393 \pm 0.008$} \\
MedAE  & \ApplyAbsoluteGradient{2}{$0.394 \pm 0.003$} 
       & \ApplyAbsoluteGradient{3}{$0.415 \pm 0.003$} 
       & \ApplyAbsoluteGradient{4}{$0.441 \pm 0.003$} 
       &  
       &  
       & \ApplyAbsoluteGradient{1}{$0.211 \pm 0.004$} \\
MSE    & \ApplyAbsoluteGradient{2}{$0.439 \pm 0.003$} 
       & \ApplyAbsoluteGradient{3}{$0.456 \pm 0.002$} 
       & \ApplyAbsoluteGradient{4}{$0.488 \pm 0.002$} 
       &  
       &  
       & \ApplyAbsoluteGradient{1}{$0.396 \pm 0.005$} \\
RMSE   & \ApplyAbsoluteGradient{2}{$0.662 \pm 0.002$} 
       & \ApplyAbsoluteGradient{3}{$0.675 \pm 0.001$} 
       & \ApplyAbsoluteGradient{4}{$0.698 \pm 0.002$} 
       &  
       &  
       & \ApplyAbsoluteGradient{1}{$0.629 \pm 0.004$} \\
R2     & \ApplyAbsoluteGradient{2}{$0.439 \pm 0.004$} 
       & \ApplyAbsoluteGradient{3}{$0.418 \pm 0.003$} 
       & \ApplyAbsoluteGradient{4}{$0.376 \pm 0.004$} 
       &  
       &  
       & \ApplyAbsoluteGradient{1}{$0.494 \pm 0.007$} \\
NLL    & \ApplyAbsoluteGradient{2}{$1.531 \pm 0.001$} 
       & \ApplyAbsoluteGradient{3}{$1.556 \pm 0.002$} 
       & \ApplyAbsoluteGradient{1}{$1.457 \pm 0.002$} 
       &  
       &  
       & \ApplyAbsoluteGradient{4}{$1.610 \pm 0.001$} \\       

\midrule
\multicolumn{7}{l}{\textbf{Concrete Slump Test Slump}} \\
MAE    & \ApplyAbsoluteGradient{5}{$5.480 \pm 0.178$} 
       & \ApplyAbsoluteGradient{2}{$4.993 \pm 0.131$} 
       & \ApplyAbsoluteGradient{1}{$4.221 \pm 0.135$} 
       & \ApplyAbsoluteGradient{6}{$5.574 \pm 0.150$} 
       & \ApplyAbsoluteGradient{4}{$5.086 \pm 0.235$} 
       & \ApplyAbsoluteGradient{3}{$5.063 \pm 0.114$} \\
MedAE  & \ApplyAbsoluteGradient{5}{$4.253 \pm 0.349$} 
       & \ApplyAbsoluteGradient{4}{$3.704 \pm 0.255$} 
       & \ApplyAbsoluteGradient{1}{$3.044 \pm 0.183$} 
       & \ApplyAbsoluteGradient{6}{$4.373 \pm 0.193$} 
       & \ApplyAbsoluteGradient{2}{$3.453 \pm 0.349$} 
       & \ApplyAbsoluteGradient{3}{$3.688 \pm 0.204$} \\
MSE    & \ApplyAbsoluteGradient{5}{$52.397 \pm 3.007$} 
       & \ApplyAbsoluteGradient{2}{$46.440 \pm 2.638$} 
       & \ApplyAbsoluteGradient{1}{$33.924 \pm 3.504$} 
       & \ApplyAbsoluteGradient{6}{$54.221 \pm 2.582$} 
       & \ApplyAbsoluteGradient{4}{$53.774 \pm 3.901$} 
       & \ApplyAbsoluteGradient{3}{$48.683 \pm 1.638$} \\
RMSE   & \ApplyAbsoluteGradient{5}{$7.025 \pm 0.208$} 
       & \ApplyAbsoluteGradient{2}{$6.612 \pm 0.158$} 
       & \ApplyAbsoluteGradient{1}{$5.603 \pm 0.239$} 
       & \ApplyAbsoluteGradient{6}{$7.142 \pm 0.225$} 
       & \ApplyAbsoluteGradient{4}{$7.017 \pm 0.315$} 
       & \ApplyAbsoluteGradient{3}{$6.748 \pm 0.128$} \\
R2     & \ApplyAbsoluteGradient{5}{$0.036 \pm 0.154$} 
       & \ApplyAbsoluteGradient{2}{$0.177 \pm 0.158$} 
       & \ApplyAbsoluteGradient{1}{$0.334 \pm 0.171$} 
       & \ApplyAbsoluteGradient{6}{$0.050 \pm 0.173$} 
       & \ApplyAbsoluteGradient{4}{$0.150 \pm 0.101$} 
       & \ApplyAbsoluteGradient{3}{$0.083 \pm 0.200$} \\
NLL    & \ApplyAbsoluteGradient{2}{$3.408 \pm 0.029$} 
       & \ApplyAbsoluteGradient{1}{$3.322 \pm 0.038$} 
       & \ApplyAbsoluteGradient{6}{$4.461 \pm 0.125$} 
       & \ApplyAbsoluteGradient{3}{$3.429 \pm 0.029$} 
       & \ApplyAbsoluteGradient{5}{$5.485 \pm 0.290$} 
       & \ApplyAbsoluteGradient{4}{$3.901 \pm 0.002$} \\

\end{longtable}
\end{landscape}

\end{appendices}



\end{document}